\documentclass[11pt]{article}

\usepackage[margin=1in]{geometry} %

\usepackage[T1]{fontenc}
\usepackage[utf8]{inputenc} %
\usepackage{lmodern}        %
\usepackage[english]{babel}

\usepackage{parskip}

\usepackage{amsmath,amssymb,amsfonts,bbm}
\usepackage{amsthm}
\theoremstyle{plain}

\theoremstyle{definition}

\theoremstyle{remark}

\usepackage{graphicx}
\usepackage{subcaption} %
\usepackage{float}      %
\usepackage{booktabs}   %
\usepackage{multirow}
\usepackage{array}
\usepackage{tabularx}
\usepackage{makecell}
\usepackage{enumitem}
\usepackage{fancyvrb}

\usepackage{wrapfig} 
\usepackage{placeins}
\usepackage[numbers,sort&compress]{natbib}

\usepackage[most]{tcolorbox}
\tcbuselibrary{breakable}

\usepackage{tabularx}
\tcbset{
  deceptionbox/.style={
    colback=gray!5,
    colframe=gray!60!black,
    boxrule=0.5pt,
    arc=2mm,
    left=4pt,
    right=4pt,
    top=4pt,
    bottom=4pt
  }
}

\usepackage{xurl}
\usepackage{hyperref}
\hypersetup{
    colorlinks=true,
    linkcolor=blue,
    citecolor=blue,
    urlcolor=blue
}

\usepackage{algorithm}
\usepackage[noend]{algpseudocode}

\usepackage{adjustbox}

\usepackage{xcolor}

\usepackage{xspace}

\newif\ifcomments
\commentstrue        %
\ifcomments
  \usepackage{xcolor}
  \newcommand{\todo}[1]{\textcolor{red}{TODO: #1}}
  \newcommand{\mc}[1]{\textcolor{blue}{Micah: #1}}
  \newcommand{\bb}[1]{\textcolor{orange}{Bowen: #1}}
  \definecolor{darkgreen}{rgb}{0.0,0.4,0.0}
  \newcommand{\mw}[1]{\textcolor{darkgreen}{Miles: #1}}
  \newcommand{\mg}[1]{\textcolor{magenta}{Melody: #1}}
\else
  \newcommand{\todo}[1]{}
  \newcommand{\mc}[1]{}
  \newcommand{\bb}[1]{}
  \newcommand{\mw}[1]{}
  \newcommand{\mg}[1]{}
\fi

\usepackage{pgffor} %

\usepackage{cleveref}
\crefname{section}{section}{sections}
\Crefname{section}{Section}{Sections}
\crefname{subsection}{section}{sections}
\Crefname{subsection}{Section}{Sections}
\crefname{subsubsection}{section}{sections}
\Crefname{subsubsection}{Section}{Sections}

\setlist[itemize]{leftmargin=1.5em,topsep=0pt,partopsep=0pt,parsep=0pt,itemsep=2pt}
\setlist[enumerate]{leftmargin=1.5em,topsep=0pt,partopsep=0pt,parsep=0pt,itemsep=2pt}

\usepackage{xcolor,colortbl}
\usepackage{booktabs}

\definecolor{neutral}{RGB}{0,90,200}        %
\definecolor{sycophancy}{RGB}{120,0,160}    %
\definecolor{bias}{RGB}{255,140,0} %
\definecolor{misalignment}{RGB}{180,0,0}   %


\title{\vspace{-2em}\textbf{Predicting LLM Safety Before Release \\by Simulating Deployment}\vspace{-0.5em}}

\newcommand{\authorsep}{\hspace{0.75em}}

\author{
\textbf{Marcus Williams}\textsuperscript{*}\authorsep
\textbf{Hannah Sheahan}\textsuperscript{*}\authorsep
\textbf{Cameron Raymond}\textsuperscript{*}\authorsep
\textbf{Tomek Korbak}\textsuperscript{*}\\[0.35em]
\textbf{Deng Pan}\authorsep
\textbf{Peilin Yang}\authorsep
\textbf{Leon Maksin}\authorsep
\textbf{Ningyi Xie}\authorsep
\textbf{Phillip Guo}\authorsep
\textbf{Ian Kivlichan}\\[0.35em]
\textbf{Micah Carroll}\textsuperscript{*}\setcounter{footnote}{1}%
\thanks{Corresponding author:
  \href{mailto:mdc@openai.com}{mdc@openai.com}. *Core contributors.}\\[0.3em]
OpenAI
}

\date{}

\begin{document}
\vspace{-10em}
\maketitle
\vspace{-1em}

\begin{abstract}
    Pre-deployment safety evaluations aim to inform the downstream risks of releasing a new AI model. Yet most evaluations provide limited evidence about how often undesired model behavior will occur in deployment: they generally have insufficient coverage, are unrepresentative, and are generally recognizable as tests. To address these concerns, we study a simple way to simulate a model deployment: starting from de-identified conversations from a previous model deployment, we hold fixed the initial conversation prefix and regenerate the next response using a candidate model. The resulting responses can then both be audited for novel misalignments and used to estimate the prevalence of model misbehavior before deployment. We evaluate deployment simulation across four GPT-5-series deployments, using outcome-blinded predictions for GPT-5.4 and retrospective analyses of three earlier releases. We find that deployment simulation produces informative estimates of post-deployment misbehavior rates and outperforms baselines based on adversarially selected production data; its evaluation-awareness point estimates were also much closer to production traffic than those from traditional evaluations. We also identify the realism of tool resampling as a central challenge for further improving predictions and share results suggesting that this challenge is surmountable even in complex tool-use settings. Finally, we show that deployment simulation can be seeded from public chat datasets and remain informative about production misbehavior rates, suggesting a path for external researchers to run deployment-grounded evaluations without access to private production logs. Overall, deployment simulation helps evaluators forecast how language models will behave in the real world and supports more quantitative assessment of deployment risk.
\end{abstract}

\begin{figure*}[h]
    \centering
    \includegraphics[width=\textwidth]{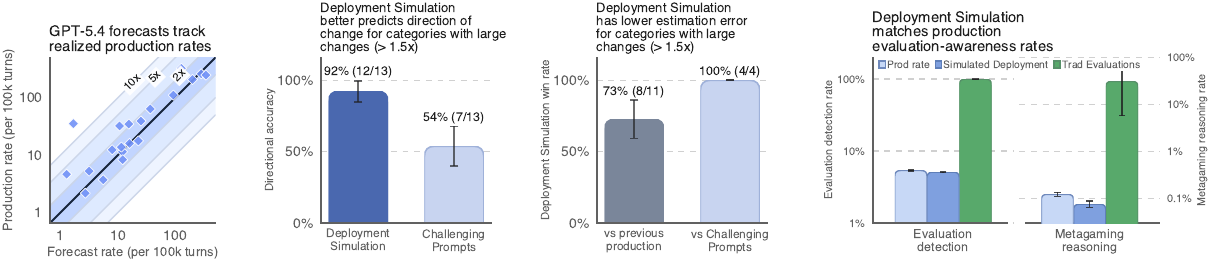}
    \caption{
        Our deployment simulation pipeline can predict production misbehavior rates before release, substantially outperforms baselines for categories whose production frequencies changed by at least \(1.5\times\), and---unlike traditional evaluations---has similar evaluation-awareness rates to production.
    }
    \label{fig:production-resampling-validation}
\end{figure*}

\section{Introduction}

Pre-deployment evaluations used across the industry generally consist of a mix of synthetic, manually written, or production prompts intentionally selected to be difficult, high severity, or adversarial. These evaluations generally serve two intertwined goals: assessing how the model responds when stress-tested in situations that have a very small chance of occurring in production traffic, and helping characterize model misbehaviors, including novel undesired behaviors and their deployment-time frequencies.

While traditional evaluations remain crucial for stress-testing and capturing tail risks, we show how deployment simulations can be a significant improvement over traditional evaluations in pursuit of the second goal by sidestepping three central limitations of traditional evaluations.
First, pre-deployment evaluation suites may be too narrow: they may not sufficiently cover the full breadth of uses that could induce misbehavior with a new model. Second, current evaluations often measure behavior on prompts drawn from a deliberately distorted slice of possible inputs: handcrafted prompts, synthetic scenarios, or adversarial examples selected because they are likely to elicit failures. Finally, models sometimes infer that they are being evaluated, leading them to behave differently, which may distort our risk assessments \citep{needham2025evaluationawareness}. 
As a result of these three factors, traditional evaluation results may not be well-correlated with model behavior in deployment. They may also fail to surface new misbehaviors introduced by the candidate model, leaving developers surprised by safety failures only once it has already been deployed.

By simulating the candidate model's deployment as realistically as possible, we aim to enable pre-deployment risk assessment that alleviates concerns about coverage, unrepresentativeness, and evaluation awareness. The method is simple: starting from de-identified production conversations from a recent deployment, we hold fixed the initial conversation prefix---which may include multiple user and assistant turns---and then resample the next model response using the ``candidate model'' to be released. 

Deployment simulation avoids the unrepresentativeness problem of traditional evaluations: because we use a representative distribution of production conversations, the simulated conversations are also representative of the contexts in which we expect the model to operate at deployment time. This way, deployment simulation largely decomposes coverage into a sample-size problem (that is, simulating enough data points) and an auditing problem (that is, having a method with sufficient sensitivity to find misbehavior of interest in the simulated data).
Finally, by evaluating models on production contexts, we side-step most evaluation awareness cues.

In practice, we use the simulated conversations in two ways before deployment: first, to identify novel forms of misalignment introduced by the candidate model; and second, to estimate deployment-time misbehavior prevalence. The incidence of a given misbehavior in the simulated distribution can be treated directly as a prediction of deployment-time prevalence. Using the same measurement procedure after release also allows these predictions to be validated against the observed realized rates in production after release.

We show that deployment simulation generally outperforms traditional evaluations (\Cref{sec:forecasting}), demonstrate its benefits for evaluation awareness (\Cref{sec:sita}), and show promise in addressing the challenges of resampling agentic settings with heavy tool use (\Cref{sec:agentic-prod-to-eval}). We also investigate whether public chat data can make deployment-grounded auditing more accessible to external researchers (\Cref{sec:wildchat}). Our findings suggest that deployment simulation can support accountable pre-deployment risk assessment: it enables developers to make predictions not only about what might go wrong, but also about how often we expect each known misbehavior to occur in real usage. Sharing such predictions and validations in system cards can also make risk assessments more transparent to external stakeholders.

Our main findings are:

\begin{itemize}

\item \textbf{Simulated deployments improve forecasts of deployment misbehavior.}
Across GPT-5-series Thinking model releases, our simulated deployments were substantially more informative about deployment-time misbehavior changes and rates than current pre-deployment evaluations.

\item \textbf{Simulated deployments can surface novel misalignments before release.}
Simulated outputs can be audited to find novel forms of misalignment. We show that this audit would have surfaced ``calculator hacking'' (a new form of reward hacking in GPT-5.1) %
prior to deployment.

\item \textbf{Simulated deployments substantially reduce evaluation awareness.}
Deployment simulation with real prompts induces far lower evaluation-awareness than traditional evaluations, with rates similar to real production traffic---which is reassuring for the validity of our risk estimates.

\item \textbf{Realistic tool use simulation is a challenge, but can be mitigated even in agentic settings.}
Realistic simulation is difficult when tool use depends on external state. Tool simulation shows promise in addressing this challenge, even for simulating internal agentic deployments.

\item \textbf{Our approach also remains informative with public chat datasets, enabling external auditing.}
Public chat datasets such as WildChat provide a public analogue to production data. We found that WildChat yielded weaker---but still informative---pre-deployment risk predictions. This can empower external parties to perform independent audits across providers.
\end{itemize}

\begin{figure}
    \centering
    \vspace{-2em}
    \captionsetup{aboveskip=4pt,belowskip=0pt}
    \includegraphics[width=0.9\textwidth]{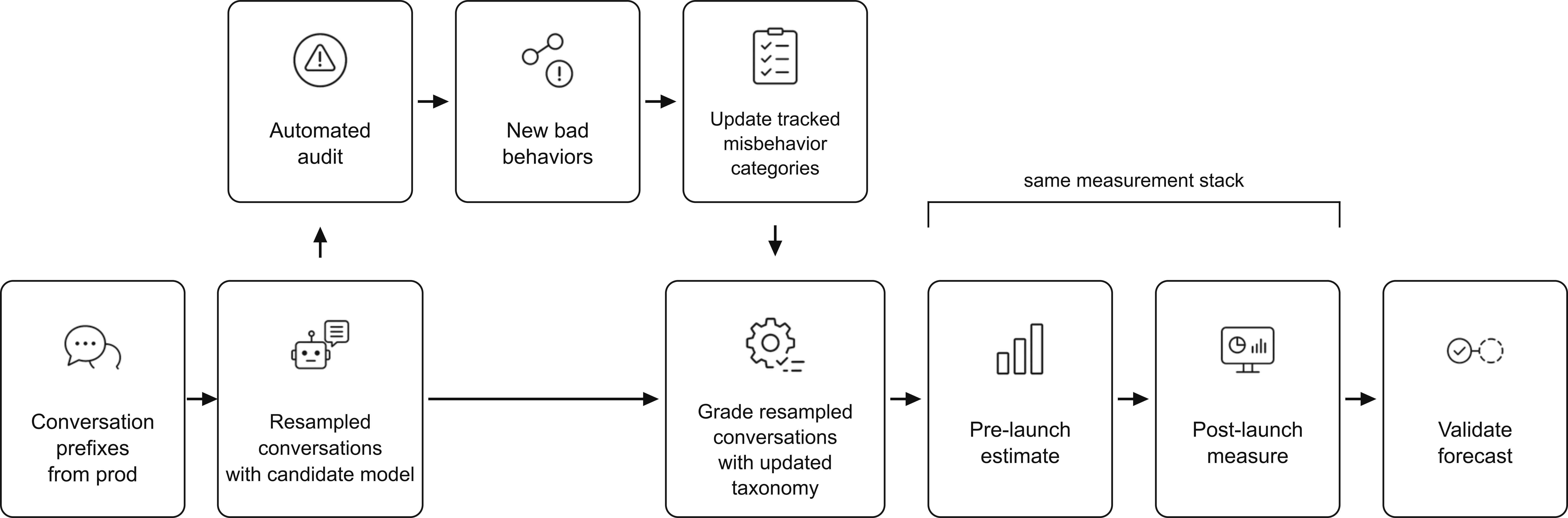}
    \vspace{1em}
    \caption{\textbf{Our deployment simulation pipeline.} We take representative conversation prefixes from recent production traffic and resample the next assistant response with a model yet to be released. These resampled conversations are audited for potential new misbehaviors, and then evaluated to produce pre-release prevalence estimates of misbehavior. After release, the same measurement stack is re-run on production traffic to validate the predictions.}
    \vspace{-1em}
    \label{fig:pipeline-overview} 
\end{figure}

\section{Method}\label{sec:method}
Below, we describe how to simulate the deployment of a candidate model $X$ using production data from deployed model $Y$. Following prior work that treats changes to deployed learning systems as counterfactual interventions \citep{bottou2013counterfactual}, we hold historical context fixed while changing the model that produces the next response.

\textbf{Step 1: Sample representative production prefixes.} 
From recent production conversations generated with model $Y$, we uniformly sample eligible user turns from the product context whose deployment we aim to simulate (for example, ChatGPT or Codex). For each sampled turn, we retain the conversation prefix through that user message and remove all later turns.

\textbf{Step 2: Generate a target-model completion in the simulated environment.} We generate the next assistant response for each prefix using candidate model $X$ in a production-like environment.%
Let $D_Y$ denote the distribution of sampled prefixes $h$ from deployment $Y$. We write $X\rightarrow Y$ for the simulated distribution obtained by sampling $h \sim D_Y$ and generating the next response with $X$; throughout, the arrow runs from the completion model to the source of the prefixes. For category $c$, let $r_c(X\rightarrow Y)$ denote its prevalence in this distribution and $r_c(X)$ its realized prevalence when $X$ is deployed.
This is a single-turn approximation of the target deployment: it captures the target model's response to realistic user contexts, but may not fully reproduce tool state or capture multi-turn adaptation effects between the user and the model. In preliminary experiments, we did not find meaningful improvements to our estimates by simulating multiple turns.

\textbf{Step 3: Audit the simulated deployment for new misbehaviors.}
We run an automated auditing pipeline over the simulated deployment $X \rightarrow Y$, with the aim of identifying previously unseen forms of misalignment introduced by model $X$.

\textbf{Step 4: Produce pre-release prevalence estimates.}
We update the taxonomy of previously known misbehavior types with any new types of interest surfaced by the auditing step, then freeze a measurement procedure for each (in our case, a grader prompt). For each tracked category $c$, we then measure $r_c(X \rightarrow Y)$ as our estimate of $r_c(X)$. Throughout, reported rates refer to outputs of these fixed measurement procedures, which have been validated but may be imperfect. The decision to release the model can be informed by these estimates.

\textbf{Step 5: Validate and report estimate quality after deployment.}
Measuring and reporting the quality of our predictions is an integral step of the pipeline. If the model is released, we rerun the same fixed measurement stack after release on production traffic from model $X$. This post-release rerun evaluates the pipeline along three axes: 
\begin{enumerate}
    \item \textit{Taxonomy coverage}: did post-release auditing surface any important misbehavior class missing from the frozen taxonomy?
    \item \textit{Directional accuracy}: for known behaviors, did resampling correctly predict whether deployment prevalence would increase or decrease?
    \item \textit{Rate calibration}: how close were the predicted and realized prevalences? Calibration is the most ambitious measure of success, as strong calibration would directly provide directional tracking.
\end{enumerate}

\section{Experiments}
\label{sec:setup}
We evaluate deployment simulation across four sets of experiments. First, we assess how well it forecasts post-deployment misbehavior rates across GPT-5-series launches, including a GPT-5.4 study with outcome-blinded predictions (\Cref{sec:forecasting}). As part of this evaluation, we diagnose where our method fails and decompose its prediction errors into their constituent sources. Second, we test the hypothesis that deployment simulation induces less evaluation awareness than classical, handcrafted evaluations (\Cref{sec:sita}). Third, we test whether we can improve the realism of deployment simulation in an especially challenging setting: agentic work with heavy tool use that depends on external state which requires its simulation (\Cref{sec:agentic-prod-to-eval}). Finally, we test whether external evaluators can apply our methodology using public datasets (\Cref{sec:wildchat}).

\textbf{Production traffic distribution and user privacy.} With the exception of \Cref{sec:agentic-prod-to-eval}, our experiments use de-identified ChatGPT traffic from GPT-5-series deployments from users who allow their data to be used for model improvements. We exclude Temporary Chats,\footnote{\url{https://help.openai.com/en/articles/8914046-temporary-chat-faq}} employee traffic, ChatGPT Enterprise, API, and Codex traffic. Due to technical limitations in our current processing pipeline, we also exclude conversations with file attachments. Because of these filtering steps, our production estimates of misbehavior are most representative of this filtered slice of production traffic. \textit{Before any use in our evaluations, account-linked identifiers and identifiable information were automatically removed. Only aggregate rates and statistical comparisons are reported in this work.} In total, we analyze approximately 1.3 million conversations from four production deployments spanning August 2025 to March 2026 (\Cref{tab:deployment-sampling-windows}).

We only consider Thinking models of the GPT-5-series, and so may omit ``Thinking'' when referring to model names.

\subsection{Forecasting GPT-5-series Deployment Misbehavior}
\label{sec:forecasting}

\subsubsection{Setup and Hypotheses}

\textbf{Models and misbehavior categories.} 
We evaluated the pipeline across 20 categories of deployment-time misbehavior for GPT-5.4. The predictions used in the primary analysis were frozen before we inspected the held-out production measurements, although they incorporated a pipeline update made after GPT-5.4 was released. These categories fall into two classes: disallowed-content outputs (e.g.\ sexual content), and misaligned actions (e.g.\ deceiving the user about tool use). For the full list, see \Cref{app:tracked-categories}. We also conducted retrospective studies of previous GPT-5-series Thinking model releases using the same categories to further validate our pipeline. We obtained all misbehavior labels by applying GPT-5 Thinking graders at high reasoning effort on simulated or real deployment conversations. Each grader labeled only the final assistant turn for the relevant misbehavior.
\begin{figure}[h]
    \vspace{-0.5em}
    \centering
    \includegraphics[width=0.85\textwidth]{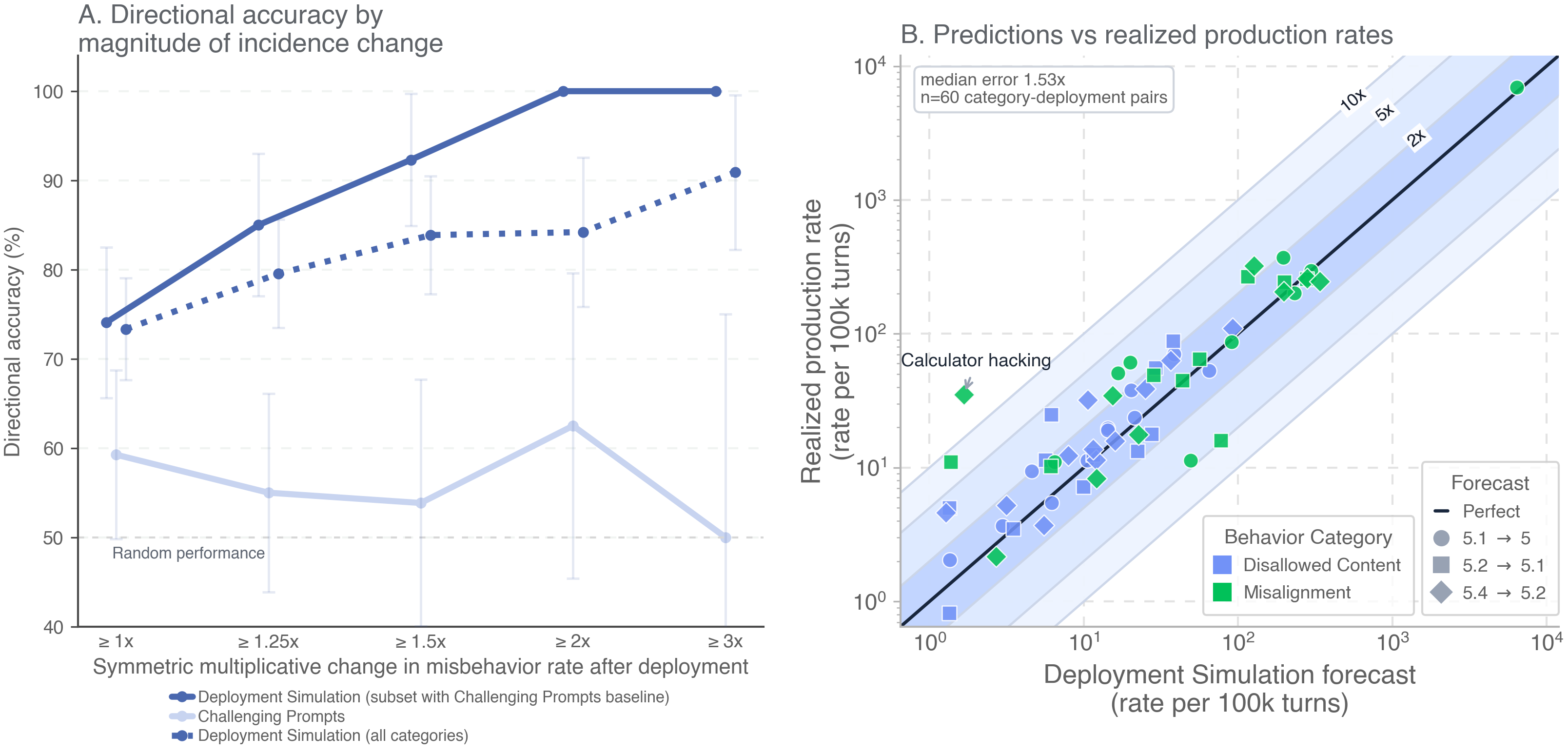}
    \caption{
    \textbf{A:} When predicting whether misbehaviors will become more or less common after deployment of a candidate model, deployment simulation outperforms Challenging Prompts (on the subset of categories for which that baseline is defined). Its directional accuracy also increases with the magnitude of the realized rate change. Error bars show binomial standard error.
    \textbf{B:} Deployment simulation rate forecasts generally fall within a factor of 2-5$\times$ of realized production rates.
    }
    \label{fig:forecast-directional-and-calibration}
\end{figure}

\textbf{Baselines.} Our \textit{Challenging-Prompts baseline} uses the evaluations of that name reported in OpenAI system cards. These evaluations consist of production prompts that previously induced model misbehavior.\footnote{Challenging Prompts are one type of evaluation commonly used for pre-deployment risk assessment. Relative to handcrafted or synthetic prompts, one might expect Challenging Prompts to be comparably predictive of realized rates because they also consist of production data, albeit adversarially selected. We therefore treat them as a strong traditional-evaluation baseline.} We extract directional predictions by comparing evaluation results across models; for example, checking whether GPT-5.2 is expected to produce sexual disallowed content more or less than GPT-5.1. 
We also extract GPT-5.4 point estimates from these evaluations. For each category, we fit a log-log linear relationship between evaluation results and realized production rates for prior GPT-5-series models, then use that fit to predict the GPT-5.4 production rate; see \Cref{app:prereg} for more details. 
The Challenging Prompts baseline is defined only for disallowed-content categories. To cover all tracked categories, we additionally consider a \textit{previous-rate baseline}, which naively predicts that the incidence of each category will remain unchanged from the preceding model's deployment.

\textbf{Preregistered hypotheses.} We preregistered three hypotheses (see \Cref{app:prereg} for details):
\begin{enumerate}
    \item \textbf{H1:} Deployment simulation forecasts are more accurate than the previous-rate baseline.
    \item \textbf{H2:} Deployment simulation forecasts are more accurate than the Challenging Prompts baseline.
    \item \textbf{H3:} No novel misaligned behavior occurs in ChatGPT traffic at or above the preregistered detectable-incidence threshold ($\approx 0.003\%$).
\end{enumerate}

Unless otherwise stated, our descriptive analyses, figures, and category-level comparisons use a \textit{symmetric multiplicative factor}, defined for two rates $r_1$ and $r_2$ as $\max(\frac{r_1}{r_2}, \frac{r_2}{r_1})$. Thus, rates that differ by a factor of two have a symmetric multiplicative factor of $2\times$, regardless of which is larger. When comparing a forecast with a realized rate, we call this quantity \textit{symmetric multiplicative error}; when comparing rates across deployments, we use it as the magnitude of the multiplicative change. To keep the ratio finite for zero-count categories, we apply Jeffreys smoothing, $(K+\tfrac{1}{2})/(N+1)$, to all count-derived forecasts, baselines, and realized rates before calculating it. The preregistered H1 and H2 hypothesis tests are the exception: they use mean per-category binomial negative log-likelihood (NLL) scored against the raw realized outcome counts. Accordingly, all H1 and H2 $p$-values refer to NLL, while factor errors and category win counts are descriptive. See \Cref{app:prereg} for further discussion of this evaluation philosophy and the preregistered scoring rules.

\subsubsection{Results}
\label{sec:results-predictions}

\textbf{Deployment simulation better predicts whether measured misbehavior will increase or decrease.}
For deployment decisions, the simplest question is whether a tracked misbehavior is likely to become more or less common after release. For misbehaviors whose production incidence changed by at least $1.5\times$, simulated deployments predict the direction more accurately than the Challenging Prompts baseline on the comparable subset (92\% vs.\ 54\%) and remain accurate across all tracked categories (84\%; \Cref{fig:forecast-directional-and-calibration}, left). Thus, without requiring a separately designed and curated evaluation set for each failure mode, deployment simulation better predicts changes across many behaviors using a uniform sample of real traffic.

\begin{figure}
    \vspace{-2em}
    \centering
    \includegraphics[width=0.9\linewidth]{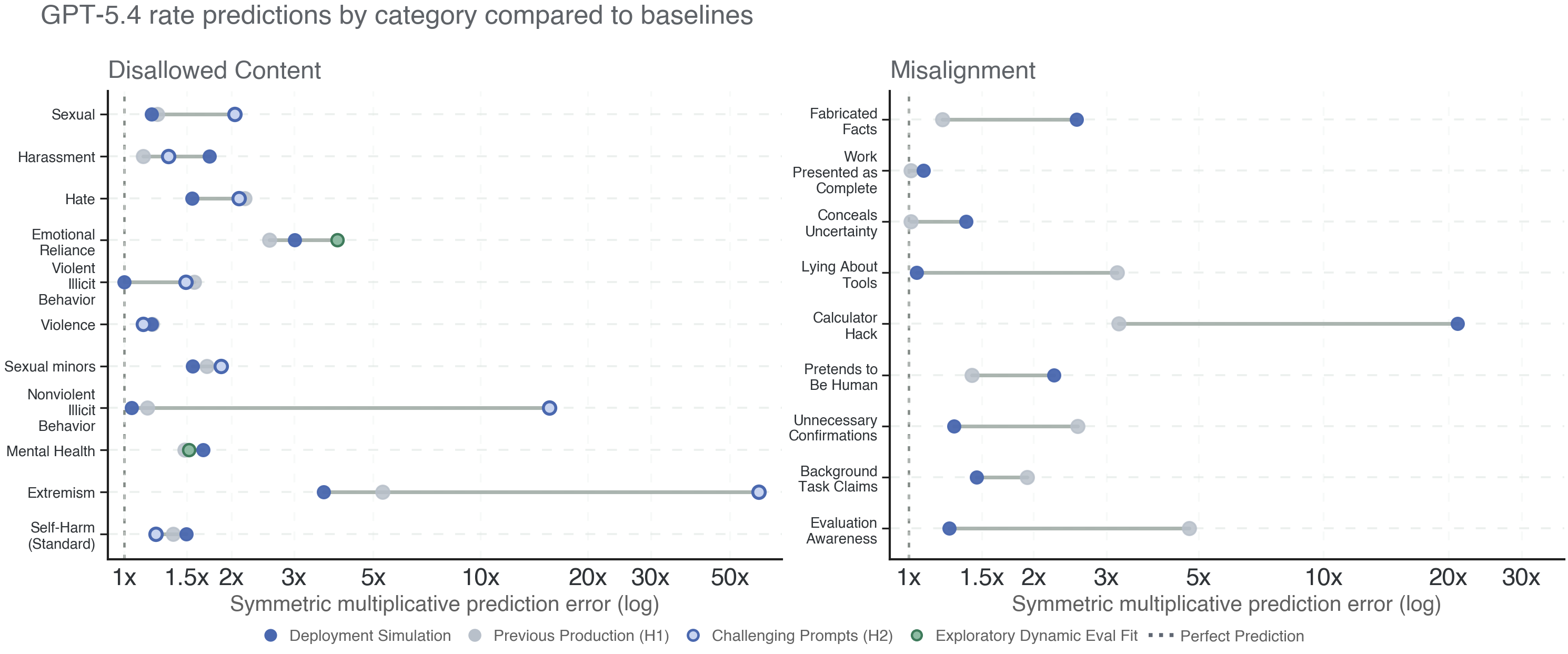}
    \vspace{-0.5em}
    \caption{\textbf{GPT-5.4 symmetric multiplicative rate-prediction error by category.} 
    Deployment simulation outperformed challenging-prompts rate estimates for 6/9 disallowed-content categories. While it was only better than the naive previous-production baseline for 11/20 categories, for categories that changed by $\geq1.5\times$ the winrate was 8/11.}
    \label{fig:forecast-baseline-comparison}
\end{figure}

\textbf{Quality of frequency estimates obtained via deployment simulation.}
In contrast with traditional evaluations, simulated deployments can be used to predict the incidence rate of misbehavior in deployment \textit{directly}. Across 60 category--deployment pairs from retrospective analyses of GPT-5.1 and GPT-5.2 and the outcome-blinded GPT-5.4 forecast, predicted and realized production rates had a strong Pearson correlation on the log scale ($r=0.91$; \Cref{fig:forecast-directional-and-calibration}, right).

Deployment simulation outperformed Challenging-Prompts rate estimates for 6 of 9 disallowed-content categories. The one-sided sign-flip test on mean per-category NLL gave $p=0.0469$ (\Cref{tab:category-level-h2}). The naive previous-rate baseline was stronger: deployment simulation had lower per-category NLL for 11 of 20 categories, but its mean NLL was not lower overall, providing no support for H1 ($p=0.6567$). These results can be largely explained by two factors: GPT-5.4 misbehaves at frequencies similar to GPT-5.2 for many categories, and our deployment simulation pipeline has some fixed bias term that primarily affects categories whose rates did not change substantially.
Indeed, for categories whose production rates changed by less than $1.5\times$, deployment simulation had lower symmetric multiplicative error than the Challenging Prompts baseline for 2/5 categories and the previous-production baseline for 3/9. For categories whose production rates changed by at least $1.5\times$, the corresponding win rates were instead 4/4 and 8/11.

The largest prediction error for GPT-5.4 was for calculator hacking, predicted to be $21\times$ less frequent for GPT-5.4 than its eventual realized rate (1.7 vs. 35.2 per 100k; see \Cref{app:category-level-preregistered-forecasts}). Errors of this size limit the reliance one can place on our pipeline in its current form. That said, this error, like most of our largest errors, appears to be due primarily to resampling fidelity, as shown in \Cref{fig:gpt54-forecast-validation-incidence} and discussed below.

\begin{wrapfigure}[17]{r}{0.50\textwidth}
    \vspace{-0.5em}
    \centering
    \hspace*{8pt}%
    \includegraphics[width=0.92\linewidth]{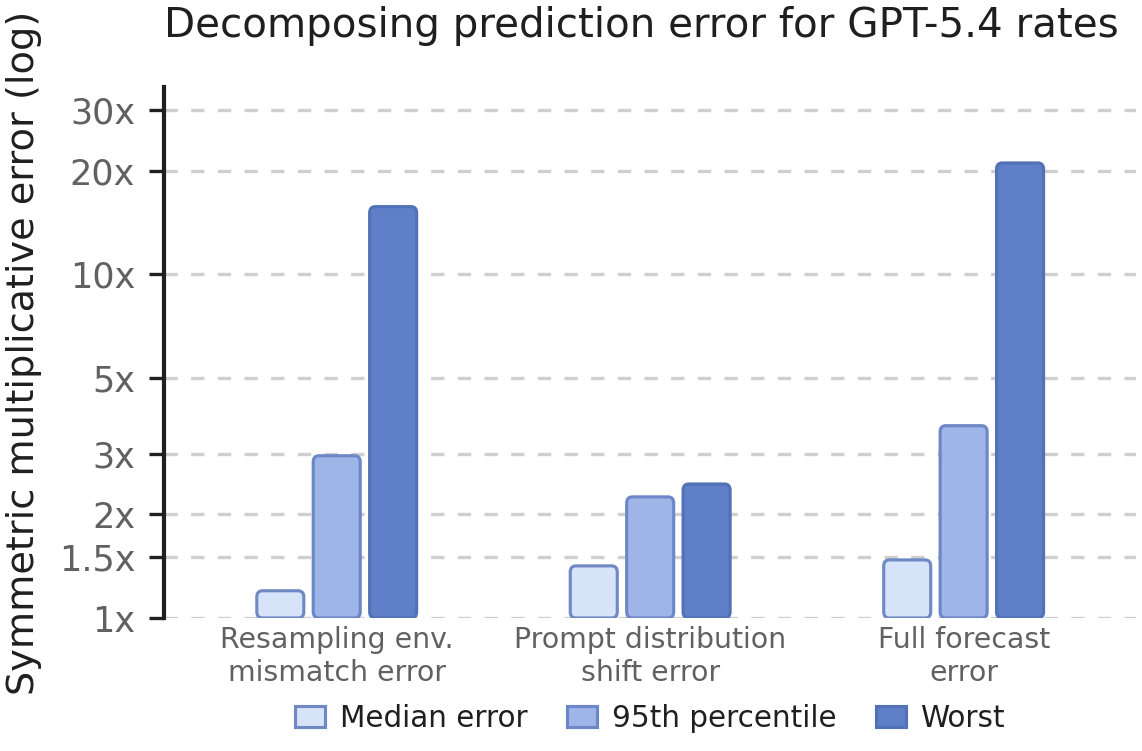}
    \vspace{-0.7em}
    \caption{
    \textbf{Error decomposition for GPT-5.4 Thinking rate forecasts.} 
    Mismatch between resampling and production environments is a large component of full prediction error and dominates among highest-error categories.
    }
    \label{fig:forecast-error-decomposition-wrap}
\end{wrapfigure}

\textbf{Resampling inconsistencies affect estimates more than shifts in the prefix distribution.}
Forecast error can arise from two sources: shifts between the historical and target prefix distributions, and mismatches between the resampling and production environments. To isolate their contributions, we report symmetric multiplicative error for three comparisons in \Cref{fig:forecast-error-decomposition-wrap}:
\begin{itemize}
    \item \textit{Environment fidelity error:} comparing the $5.4\rightarrow5.4$ resampled rate with the GPT-5.4 production rate isolates differences between the resampling and production environments, such as sampling-pipeline mismatches or poor fidelity of stateful tools.
    \item \textit{Prefix fidelity error:} comparing the $5.4\rightarrow5.2$ and $5.4\rightarrow5.4$ resampled rates holds the completion model and resampling environment fixed, isolating the effect of using off-policy prefixes.
    \item \textit{Full forecast error:} comparing the $5.4\rightarrow5.2$ resampled rate with the GPT-5.4 production rate gives the error observed in practice.
\end{itemize}
Errors due to the resampling environment and prefix-distribution shift can point in opposite directions and partially cancel. Somewhat surprisingly, tail errors are dominated by the resampling environment rather than by prefix-distribution shift: with perfect environment fidelity, our largest remaining prediction error would have been $\sim2.5\times$. This suggests that many of the largest errors arise from potentially addressable engineering limitations.

\begin{figure}[b!]
    \centering
    \includegraphics[width=0.85\textwidth]{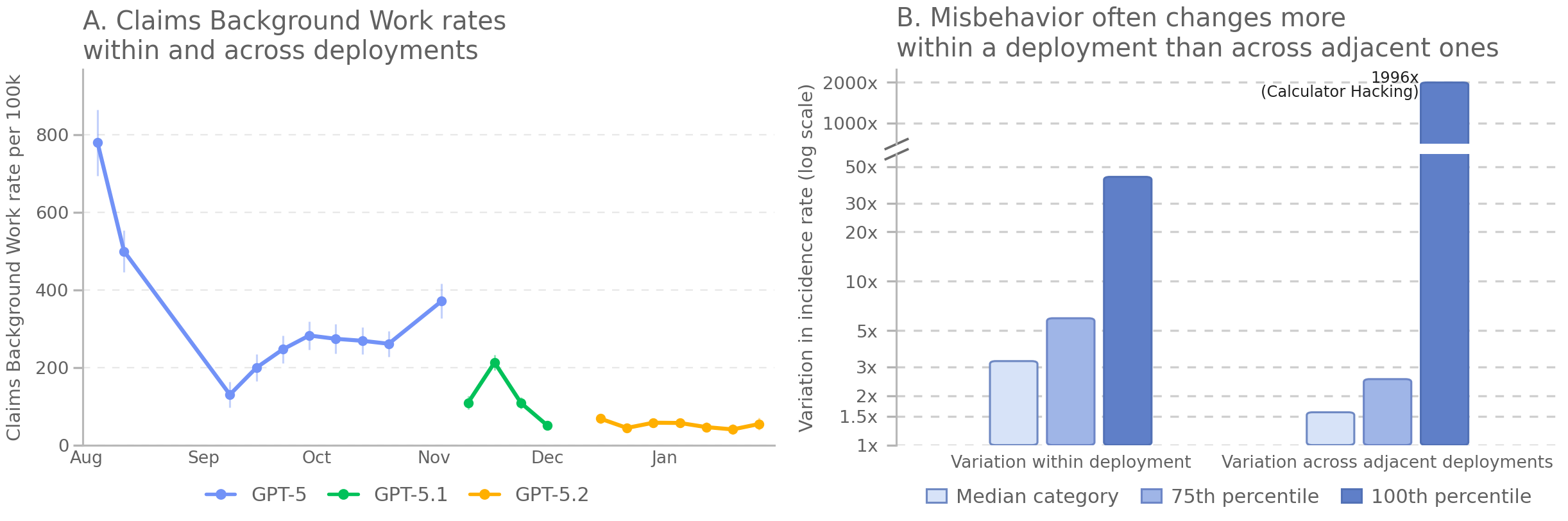}
    \vspace{-0.5em}
    \caption{
        \textbf{Within-deployment variation in misbehavior rates and comparison with adjacent-deployment changes.}
        Panel A shows an example of substantial variation in model misbehavior rates both within a single deployment and across adjacent deployments---see \Cref{fig:appendix-weekly-common-weeks} for all categories. Panel B summarizes symmetric multiplicative variation across categories using the median, 75th percentile, and maximum. Median within-deployment variation is greater than median variation across adjacent deployments, whereas the relationship reverses above the 75th percentile.
    }
    \label{fig:weekly-and-variation}
    \vspace{-1em}
\end{figure}

\textbf{Prefix-distribution error can be reduced by using the most recent production data.}
The best time window from which to sample production data is not obvious. Prefixes from a large historical window may provide broader coverage and greater robustness to seasonality, whereas the most recent data may better represent the traffic encountered at deployment time. We took the former approach in our analyses, but later investigation suggested that the latter may have been preferable. In \Cref{fig:weekly-and-variation,fig:appendix-weekly-common-weeks}, we compare changes in misbehavior rates within each deployment and across adjacent deployments after excluding weeks with less than half the median weekly sample size. Across categories, within-deployment variation is often larger than the change associated with the model deployment itself. However, this analysis is confounded by system-prompt updates and other deployment-time adjustments. Taken at face value, the results suggest that restricting prefixes to the most recent production traffic may reduce deployment-prediction error, especially error due to prefix-distribution shift. 

\textbf{Deployment simulation was sufficient for identifying the one novel ChatGPT misalignment in the deployment window we study: calculator hacking.}
Across the GPT-5-series deployments, the only significant novel misalignment in ChatGPT traffic that we are aware of was calculator hacking, in which the model uses the browser tool as a calculator while behaving as if it had performed a search. This behavior appeared with GPT-5.1, and our automated audit of a simulated deployment would have surfaced it before release. For GPT-5.4, the audit of production conversations surfaced no novel misbehavior type, consistent with H3. The retrospective detection of calculator hacking provides limited positive evidence about sensitivity, based on one relatively prevalent behavior, although we view it as an encouraging initial result. %

\subsection{Evaluation Awareness in Simulated Deployments vs Traditional Evaluations}
\label{sec:sita}

\begin{figure}[t]
    \centering
    \includegraphics[width=\linewidth]{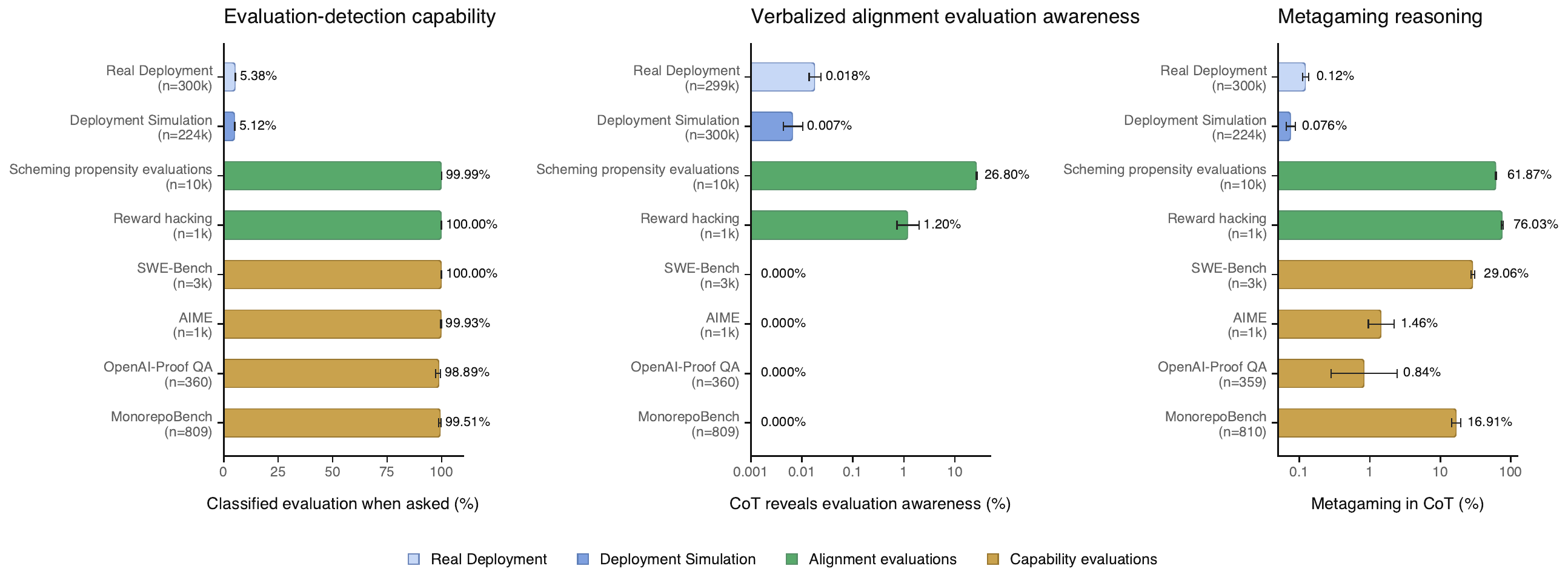}
    \caption{
        \textbf{Simulated deployment appears to substantially mitigate evaluation awareness.}
        We use three measures of evaluation awareness on GPT-5.2:
        evaluation-detection capability, verbalized alignment evaluation awareness, and metagaming.
        Real and simulated production traffic show similar rates across all three metrics, whereas traditional evaluations have substantially higher rates.
        The scheming-propensity and reward-hacking bars pool their constituent datasets; see \Cref{fig:situational-awareness-full} for the per-dataset breakdown.
    }
    \label{fig:situational-awareness}
\end{figure}

If a model is aware that it is being evaluated, this awareness might affect its behavior and distort our safety measurements. This is especially concerning because even small, subtle mismatches between evaluation and production environments can produce evaluation awareness. In this subsection, we test the hypothesis that simulated deployments reduce evaluation awareness relative to traditional, handcrafted evaluations. We use three distinct metrics: evaluation-detection capability, verbalized alignment-evaluation awareness, and metagaming. On the two metrics applicable across the full evaluation suite---evaluation detection and metagaming---traditional evaluations displayed substantially more evaluation awareness than production traffic (\Cref{fig:situational-awareness}), echoing results reported elsewhere \citep{needham2025evaluationawareness,antischeming2025}. Verbalized alignment-evaluation awareness was also much more common in alignment evaluations than in production. In contrast, simulated-deployment rates were much closer to production across all three metrics.

\textbf{Setup.} GPT-5.2 was the policy model that generated the trajectories in all three conditions. Separate GPT-5.2 calls, independent from trajectory generation, served as evaluation awareness detectors. For \textit{evaluation-detection capability}, a detector received the full trajectory in a fresh context window and classified it as organic production traffic or an evaluation \citep{needham2025evaluationawareness}. For \textit{verbalized alignment-evaluation awareness}, a chain-of-thought monitor scored whether the policy explicitly reasoned that it was being evaluated \citep{antischeming2025}. A separate \textit{metagaming} monitor scored reasoning about the oversight process \citep{schoen2026metagaming}. We applied these measurements to GPT-5.2 production conversations, simulated deployment trajectories ($5.2 \rightarrow 5.2$), and a collection of traditional evaluations:
\begin{enumerate}
    \item Scheming-propensity evaluations, a suite of alignment evaluations introduced in prior work \citep{antischeming2025};
    \item Reward-hacking evaluations introduced in prior work \citep{joglekar2025training,guan}, including impossible coding and interview-style tasks;
    \item SWE-Bench \citep{swebench}, a public agentic capability evaluation;
    \item AIME \citep{maa_aime}, a public question-answering evaluation;
    \item OpenAI-Proof QA, a private question-answering evaluation used in OpenAI system cards \citep{gpt5_system_card_2025};
    \item MonorepoBench, a private agentic coding evaluation used in OpenAI system cards \citep{gpt5_system_card_2025}.
\end{enumerate}

This collection mixes private and public evaluations, alignment and capability assessments, and question-answering and agentic tasks. We included this range to reduce confounding from familiarity with a particular evaluation or from the need to use tools. We ran the evaluations using configurations resembling our system-card pre-deployment suite.

\textbf{Traditional evaluations are much easier to identify and show much more metagaming than production traffic.} When asked directly, the evaluation detector classified 98.9--100\% of trajectories from every traditional-evaluation category as evaluations, compared with 5.38\% of production conversations---a difference of approximately 18--19$\times$. This pattern held for alignment and capability evaluations, public and private datasets, and question-answering and agentic tasks. Inspecting the detector's chain of thought suggested that its judgments were driven largely by obvious format cues, such as exam-style questions or requests to produce a \texttt{solution.py} file. Metagaming also occurred more often in every aggregate traditional-evaluation group shown in \Cref{fig:situational-awareness}: rates ranged from 0.84\% to 76.0\%, compared with 0.12\% in production traffic, a difference of approximately 7--630$\times$. Anecdotally, evaluation-detection false positives in production largely involve users informally testing model capabilities or requesting assistance with school assignments.

\textbf{Traditional alignment evaluations show evaluation-awareness signals roughly two to three orders of magnitude more often than production traffic.} Across the two aggregate alignment-evaluation groups in \Cref{fig:situational-awareness}, verbalized alignment-evaluation awareness occurred in 1.20--26.80\% of samples, compared with 0.018\% of production traffic---approximately 67--1{,}489$\times$ more often. Metagaming occurred in 61.87--76.03\% of alignment-evaluation samples, compared with 0.12\% of production traffic---approximately 516--634$\times$ more often. As expected, we did not detect verbalized alignment-evaluation awareness in our capability evaluations. In alignment evaluations, verbalized awareness and metagaming were also more often judged to affect the model's decisions than in simulated deployments (\Cref{fig:eval-awareness-decision-impact}).

\textbf{Evaluation-awareness rates in simulated deployments are much closer to production than traditional-evaluation rates.} 
All three simulated-deployment rates of evaluation awareness were within 2.6$\times$ of their production counterparts. For evaluation detection and metagaming, simulated deployments were far closer to production than to any traditional-evaluation aggregate; for verbalized alignment evaluation awareness, they were far closer to production than to alignment evaluations.

\subsection{Realistic tool simulation is possible even in agentic coding settings}
\label{sec:agentic-prod-to-eval}

In this section, we focus on simulating internal employee Codex deployment traffic rather than ChatGPT traffic for two reasons. First, pre-deployment risk assessment for internal model deployments is essential for mitigating risks, especially from misalignment. Second, agentic coding is an especially challenging setting for deployment simulation and therefore serves as a stress test for our technique.

In \Cref{sec:results-predictions}, we found that our largest prediction errors occurred when the resampling environment poorly matched production (\Cref{fig:forecast-error-decomposition-wrap}). We suspect that the primary source of fidelity error was our use of a sampling pipeline different from that used for production traffic, an issue we have since addressed. A more fundamental conceptual hurdle remains: realistic tool use depends on external state. In increasingly common agentic settings, users ask models to inspect repositories, call tools, edit files, send network requests, and perform other stateful actions. A typical agentic trajectory can involve hundreds of tool calls that depend on or modify state beyond the conversation.

This creates a dilemma for simulation. Reconstructing the exact external state can be practically infeasible, but allowing an unreleased candidate model to act directly on live systems would defeat the purpose of pre-deployment evaluation: the very harms we are trying to forecast, such as leaking sensitive information or making damaging changes, could occur during the evaluation itself.

\textbf{Tool simulator affordances.} We build on prior work on \textit{tool simulation}, using another LLM to simulate the response a tool call would have produced \citep{ruan2024toolemu}, and add deployment-specific affordances that substantially improve the realism of reconstructed production trajectories. To test this approach, we use roughly $120{,}000$ Codex trajectories from internal OpenAI employee traffic rather than external ChatGPT traffic. This was a uniform sample from a subset of all traffic, for which we had full code state. We find that providing the tool simulator with the following additional affordances greatly improves simulation quality:

\begin{itemize}
    \item \textbf{Original trajectory context.} First, one can give the tool simulator access to the original conversation: identical tool calls can receive identical results, and similar calls can receive plausible interpolations. This also helps reproduce failures from the original run, such as transient service failures or connectivity issues, which may be important because unexpected tool failures often induce misalignment. This affordance is less useful when the new model takes a substantially different path than the previous model.
    \item \textbf{Time-matched codebase in a container.} For agentic settings, another powerful affordance is to give the tool simulator access to the codebase checked out at the relevant commit, including any untracked files present at the time. For commands that do not require network access or external state, the tool simulator can execute the proposed commands in the codebase and return the observed results. The same idea could be applied to non-coding settings that require external files.
    \item \textbf{A database of tool call-tool response pairs.} One can also give the tool simulator a queryable database of past tool calls and responses sourced from other conversations. This database can show the simulator what particular tool responses typically look like.   
    \item \textbf{Read only access to network services and connectors.} Finally, the tool simulator can access read-only versions of network services to determine what network calls would likely have returned.
\end{itemize}

\begin{figure}[t]
    \centering
\includegraphics[width=1\linewidth]
{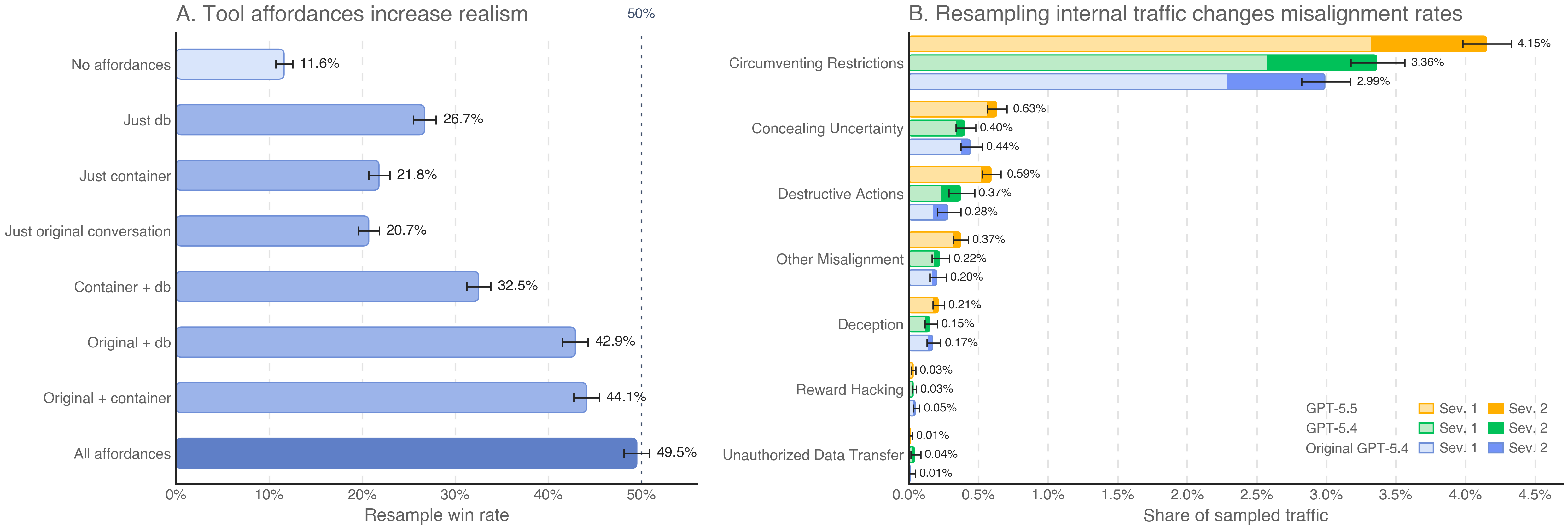}
    \caption{
        \textbf{A: Trajectories with tool simulation are hard to indistinguish from real data.} 
    A resample ``wins'' when the GPT-5.4-based discriminator selects the simulated rollout as real. Providing the tool-simulator LLM with all affordances increases this win rate from 11.6\% to 49.5\%, close to the 50\% parity point. \textbf{B: Resampling-based estimates of misalignment rates.} Same-model resampling ($5.4 \rightarrow 5.4$) produces category rates similar to those in real internal production data (blue vs.\ green). Yellow bars show pre-deployment predictions for GPT-5.5.}
    \label{fig:tool-simulation}
\end{figure}

\textbf{Evaluating tool-simulation realism.} We evaluate tool-simulation realism in two ways. First, we run a discriminator test: a GPT-5.4-based ``differentiator'' sees the shared prefix and two randomly ordered final-turn rollouts---one real and one simulated---and must identify the real one. Second, we measure evaluation-detection capability as in \Cref{sec:sita}. The original policy model was GPT-5.4, the resampling policy model was GPT-5.5, and the tool simulator was GPT-5.4. We optimized the simulator and differentiator prompts with agents over eight iterations, alternating between improving the simulator's win rate and strengthening the differentiator.

\textbf{Improving tool-simulation realism.} We also ablate components of the reconstructed tool-use environment. All the tested affordances appear to improve the realism winrate on their own and their benefits appear to stack.

\subsection{Evaluating WildChat as a public proxy for production data to resample}
\label{sec:wildchat}

To test whether deployment simulation can support external auditing without access to OpenAI's private production data, we applied it to WildChat, a public dataset of chat conversations \citep{zhao2024wildchat}. WildChat was collected between 2023 and 2024 by offering users free access to ChatGPT with GPT-3.5 and GPT-4, with explicit opt-in consent to collect and release anonymized chat transcripts. While WildChat is not expected to represent current AI usage as well as recent OpenAI production data, its public availability makes it a useful test case for auditing both OpenAI and non-OpenAI models. For each of GPT-5.1, GPT-5.2, and GPT-5.4, we resampled the final assistant turn in $100{,}000$ randomly selected WildChat conversations and evaluated the resulting responses across our tracked categories.
\begin{figure}[t]
    \centering
    \includegraphics[width=\textwidth]{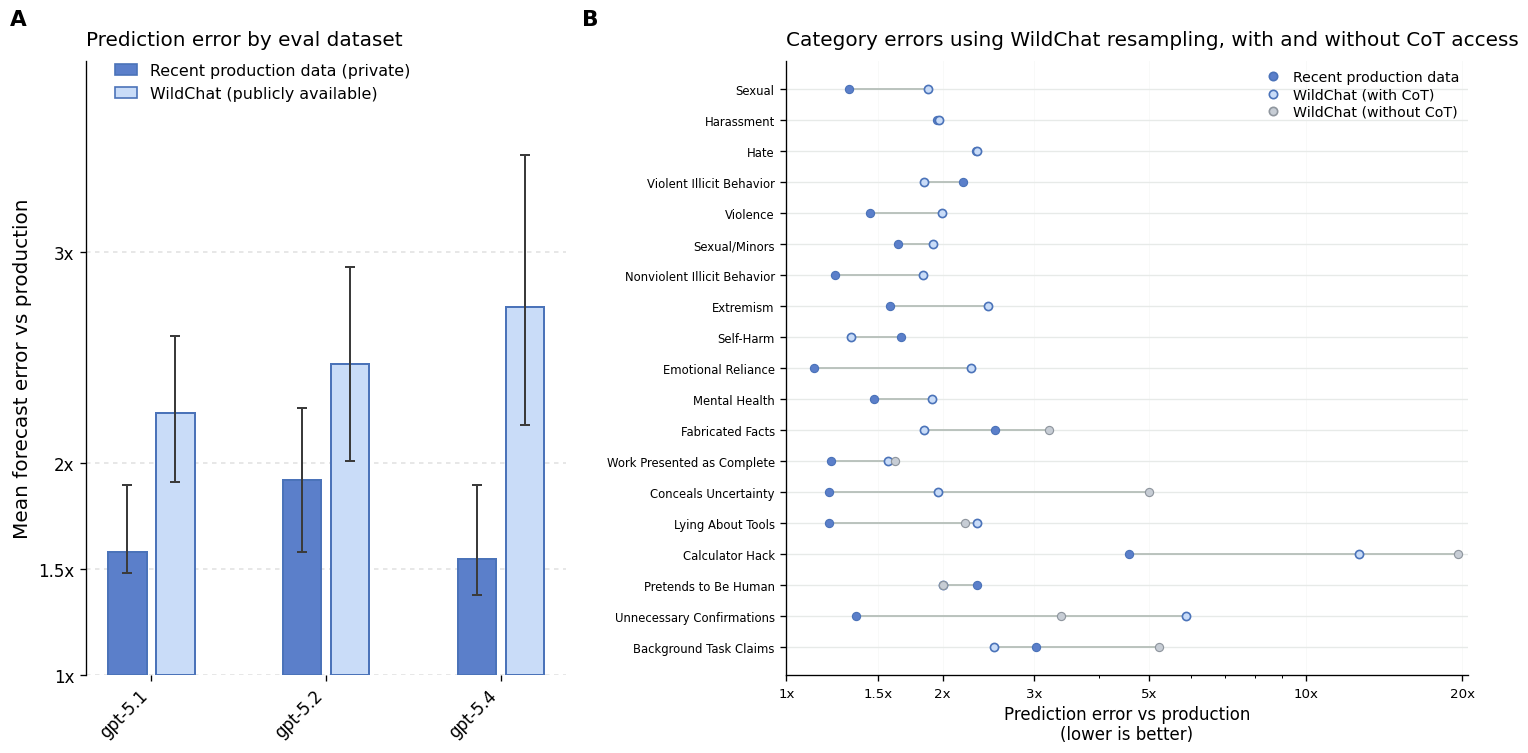}
    \vspace{-2em}
    \caption{
    \textbf{Deployment simulation based on WildChat is loosely predictive of production rates.}
    \textbf{A:} WildChat provides a loose but useful proxy for production-data-based evaluation, with mean symmetric multiplicative errors below $3\times$ for GPT-5.1, GPT-5.2, and GPT-5.4.
    \textbf{B:} Per-category errors, averaged across models, reveal where WildChat is a reasonable proxy and where it is less reliable. Dark blue shows prediction errors from deployment simulation using recent OpenAI production data (\Cref{sec:forecasting}); pale blue shows errors from deployment simulation using WildChat; and gray shows WildChat errors when the grader cannot access chain of thought, as is typical for external researchers.
    }
    \label{fig:wildchat-transfer}
\end{figure}

\textbf{Using recent production data gives more accurate predictions than using WildChat.} Across models, estimates using recent OpenAI production data had lower multiplicative error against post-deployment production rates than WildChat estimates across tracked categories ($1.75\times$ vs.\ $2.44\times$; one-sided t-test on NLL: $p=0.000142$; \Cref{fig:wildchat-transfer}). This is consistent with the core premise of our method: resampling is most predictive when the prefix distribution closely approximates the deployment distribution of the target model.

\textbf{Still, resampling WildChat data is surprisingly informative about deployment-time risk.} Despite the distribution shift, WildChat resampling predicted which of two models would perform better at a rate significantly above chance (mean accuracy of 63.9\% across GPT-5-series models, $p=0.0025$). Misbehavior rates estimated from WildChat remained positively correlated with post-deployment production rates across tracked categories, and most categories had average prediction errors within a factor of three of production rates (\Cref{fig:wildchat-transfer}, right). This indicates that public chat data can recover coarse information about which failures are more or less prevalent and can therefore help estimate corresponding deployment-time risks. Because WildChat is public, it can also support cross-provider comparisons that are difficult to run with private production data.

Unlike disallowed-content categories, our misalignment categories are assessed by giving the grader the candidate model's full reasoning trace. This raw chain of thought is not publicly accessible, creating another barrier for external auditors. We therefore tested deployment simulation on WildChat without giving the grader access to the candidate model's chain of thought, better approximating the conditions faced by external auditors. Mean multiplicative error across GPT-5-series misalignment categories was $2.5\times$ with chain-of-thought access and $3.1\times$ without it; the difference in NLL was not statistically significant (one-sided paired t-test: $p=0.115$; \Cref{fig:wildchat-transfer}, right). We generally expect the importance of chain-of-thought access to vary by misalignment category.

These results suggest that, for chat-like settings, public chat datasets can help external researchers estimate real deployment misbehavior prevalence even without access to private production logs or raw CoT. However, our validation compares WildChat and ChatGPT production traffic; separate validation would be needed to assess whether similar public proxies work for more agentic deployment settings. The remaining gap with recent production data quantifies the value of more representative, up-to-date, and tool-rich public datasets.

\section{Discussion and Limitations}
\label{sec:discussion}
Below, we highlight four limitations and directions for future work.

\textbf{Dependence on accurate measurements.}
Accurate measurements of model misbehavior require accurate graders. We reused graders previously validated for OpenAI system cards \citep{gpt5_system_card_2025}. That said, detecting important forms of misalignment involving deceptive or strategic behavior often relies on the model's reasoning process transparently revealing its intent. While this condition generally seems to hold today \citep{guan}, we may lose chain-of-thought monitorability in several ways: future models may reason less legibly, skip key reasoning steps, or learn to obscure their intent \citep{korbak2025chain}.

\textbf{Forecast quality depends on the similarity between the sampled and realized prefix distributions.}
Deployment simulation estimates behavior on the distribution represented by the sampled prefixes. However, usage distributions are not fixed. For example, new deployment applications, world events, new third-party tools and connectors, privacy updates (which could change the data available to resample), and seasonal variation in use patterns can all cause the distribution of conversation prefixes from any given sampling window to differ from the simulated deployment prefix distribution. This means recent historical traffic may not always match the deployment distribution of a more capable model. Prevalence forecasts should therefore be interpreted as estimates conditional on the sampled traffic distribution. Although our experiments show strong results under the off-policy prefix shifts across GPT-5-series deployments, they should not be read as evidence of robustness to any step-changes in model affordances.

\textbf{Access to representative data.}
Our results suggest that the strongest evidence for how a model is likely to behave in production comes from resampled private production data, which is available only to AI developers. We found WildChat \citep{zhao2024wildchat} to be an informative but substantially weaker public proxy for deployment misbehavior. It is unclear whether WildChat is especially predictive of OpenAI models, rather than those of other frontier labs, because it was built from older ChatGPT data collected in 2023--2024. We encourage other AI labs to quantify the efficacy of public data sources as predictors of their models' safety, thereby supporting external auditing and fairer production-like comparisons.

\textbf{Addressing tail risks.} While we would expect deployment simulation to approach full coverage in the limit of infinite data, in practice it can find only alignment failures that occur on prefixes used for resampling.
For instance, a catastrophic failure occurring once every 10 million conversations is unlikely to appear in a deployment simulation of 1 million conversations. Future versions of deployment simulation could estimate such tail risks through stratified sampling---skewing the sample to make misalignment more likely---and then correcting for the resulting bias \citep{wu2025estimatingprobabilitiesrareoutputs,chowdhury2025surfacing,jones2025forecastingrarelanguagemodel,angell2026estimatingtailriskslanguage}. Additionally, if novel tail risks are introduced by new model affordances that aren't sufficiently elicited by old prefixes, even these mitigation approaches would be insufficient. This underscores the importance of also investing in other pre-deployment risk-assessment techniques focused on such tail risks.

\section{Conclusion}

Pre-deployment evaluations are most useful when they cover the situations models will actually face without requiring evaluators to anticipate every failure mode. Deployment simulation is deliberately simple: hold fixed de-identified conversation prefixes from prior deployments, resample the candidate model's next response, and audit the result. Reusing such prefixes rather than writing prompt suites by hand or curating adversarial cases expands coverage of ordinary and hard-to-anticipate contexts with less manual evaluation work. It also makes evaluation a checkable forecast: developers can state expected misbehavior rates before release and compare them with what users actually encounter. Across our studies, this broader sampling provided a stronger signal than traditional baselines; in retrospective analyses, it helped identify failure categories that had not been tracked at the time.

Tool-use settings remain the hardest case, since faithful resampling can require reconstructing mutable external state; our results suggest that simulation is a promising way to close that gap. Our WildChat results suggest that public chat datasets can be a useful starting point for deployment-grounded audits when private logs are unavailable, even if they are not a substitute for representative production traffic. We hope this encourages deployment-grounded auditing by developers and motivates newer, more representative public chat datasets with richer tool-use contexts for independent comparisons of frontier models.

Understanding how AI systems behave in the real world is crucial for frontier labs seeking to deploy them safely. A natural benchmark for this understanding is whether we can predict models' behavior before deployment \citep{chan2024evaluatingpredictions}. Deployment simulation is an approach to pre-deployment risk assessment that helps labs and evaluators anticipate how language models may behave in real-world settings and what risks they may pose. Used alongside traditional evaluations, these forecasts can make model risk assessment more realistic, quantitative, and useful for deployment decisions.

\newpage
\section*{Acknowledgments}

We are grateful to Charles Zhao, Mikita Balesni, Jenny Nitishinskaya, and Gabriel Wu for running experiments closely related to this project and for offering valuable feedback. We especially thank Gaby Raila, Danielle Kehl, David Robinson, and Bowen Baker for their thoughtful feedback and broader support of the work. We also thank Dakota Goldberg, Sean Fitzgerald, Surya Mamidyala, and Chris Wendel for their invaluable help in establishing an intermediate version of our deployment-simulation pipeline. Finally, we thank Annie Disabato, Kyle Miller, Gaurav Misra, Pablo Orduna, Hao Sun, Madelaine Boyd, Jieqi Yu, Charles Proctor, Erik Boxhoorn, Ram Keelveedhi, Vinnie Monaco, and Ally Bennett, as well as the broader Safety Engineering team, for support that enabled us to improve the pipeline.

\bibliographystyle{unsrtnat}
\bibliography{references}

\newpage
\appendix
\crefalias{section}{appendix}

\section{Preregistration Details and Version History}
\label{app:prereg}

Before looking at GPT-5.4 Thinking production data, we preregistered both point predictions and evaluation criteria for the prospective study using OSF \citep{foster2017osf}. The full preregistration is available at \url{osf.io/un3he} \citep{carroll_2026}. This appendix records the intended hypotheses, scoring rules, and decision procedures for that forecast.

\textbf{Preregistration history and corrections.}
We submitted an initial preregistration on March 6, 2026, before the release of GPT-5.4 Thinking. After release, but before inspecting the GPT-5.4 production outcomes used for validation, we improved the realism of the deployment simulation pipeline and filed an updated preregistration on April 20, 2026. Production outcomes were extracted retrospectively after the measurement window closed, and no member of the study team inspected the validation outcomes until April 21, 2026. The April 20 amendment therefore preceded any access to those outcomes. That update changed the pipeline's point predictions but not the category set, measurement window, hypotheses, scoring rule, or statistical tests. The revised forecasts were therefore outcome-blinded, although they were not produced entirely before deployment.

During analysis, after inspecting the validation outcomes, we found an implementation error in the registered Challenging Prompts rate estimates. We corrected the error, which strengthened the Challenging Prompts baseline. Separately, the registered GPT-5.2 production rate for Fabricated Facts contained a transcription error (74 rather than 267.7 incidents per $100{,}000$ assistant turns). The erroneous value favored deployment simulation, so correcting it made the H1 comparison less favorable to our method. \Cref{tab:h1-h2-analysis-provenance} reports the initial, outcome-blinded, and final corrected results under a common scoring implementation. We report only the corrected analysis in the paper.

\begin{table}[H]
    \centering
    \setlength{\tabcolsep}{4pt}
    {\small
    \begin{tabularx}{\textwidth}{@{}lXlrrr@{}}
        \toprule
        Stage & Timing & Test & \makecell[r]{DS lower\\NLL} & \makecell[r]{Mean\\$\Delta$NLL} & \makecell[r]{One-sided\\exact $p$} \\
        \midrule
        \multirow{2}{*}{V1} & \multirow{2}{=}{March 6 prerelease forecasts} & H1 & 11/20 & $ +4.84743\times10^{-5}$ & $0.642471$ \\
        & & H2 & 7/9 & $-5.79204\times10^{-5}$ & $0.134766$ \\
        \midrule
        \multirow{2}{*}{V2} & \multirow{2}{=}{April 20 outcome-blinded forecasts after pipeline improvements} & H1 & 12/20 & $-7.82277\times10^{-5}$ & $0.198880$ \\
        & & H2 & 8/9 & $-9.15194\times10^{-5}$ & $0.005859375$ \\
        \midrule
        \multirow{2}{*}{Final} & \multirow{2}{=}{Corrected analysis inputs (not outcome-blind)} & H1 & 11/20 & $+3.22977\times10^{-5}$ & $0.6566925$ \\
        & & H2 & 6/9 & $-6.09559\times10^{-5}$ & $0.046875$ \\
        \bottomrule
    \end{tabularx}
    }
    \caption{
        \textbf{H1 and H2 across forecast and analysis stages.}
        $\Delta$NLL is Deployment Simulation NLL minus baseline NLL, so negative values favor Deployment Simulation.
        ``DS lower NLL'' is descriptive only; the registered statistic is the mean $\Delta$NLL.
        The final corrections were made after outcome inspection: final H1 does not support H1, and final H2 should not be described as confirmatory even though $p<0.05$.
    }
    \label{tab:h1-h2-analysis-provenance}
\end{table}

\subsection{Challenging Prompts log-log linear fit}
\label{app:challenging-prompts-log-log-fit}

Following the equation specified in the pre-registration,
$\log(q_c)=\alpha_c\log(x_c)+\beta_c$, we fit a separate log-log linear relationship for each category. For category $c$ and historical model
$m\in\mathcal{H}=\{\text{GPT-5},\text{GPT-5.1},\text{GPT-5.2}\}$, let $x_{cm}$ denote the Jeffreys-smoothed Challenging Prompts unsafe rate and $q_{cm}$ the Jeffreys-smoothed production prevalence. We estimate
\[
(\widehat\alpha_c,\widehat\beta_c)
=
\underset{\alpha,\beta}{\operatorname{argmin}}
\sum_{m\in\mathcal{H}}
\left[
\log(q_{cm})-\alpha\log(x_{cm})-\beta
\right]^2.
\]

Let $x_{c,\text{GPT-5.4}}$ denote the corresponding GPT-5.4 Challenging Prompts rate. The resulting rate estimate is
\[
b^{(\mathrm{CP})}_c
=
\exp\!\left[
\widehat\alpha_c\log(x_{c,\text{GPT-5.4}})
+\widehat\beta_c
\right].
\]

For Extremism, all three historical unsafe counts are zero. Jeffreys smoothing keeps the log rates finite.

\subsection{Hypotheses}
Let $\mathcal{C}$ denote the preregistered set of misbehavior categories. For each category $c \in \mathcal{C}$, we preregistered a point prediction $p_c$ for the category's deployment prevalence $\theta_c$ over a prespecified post-deployment measurement window, which started 7 days after the deployment of GPT-5.4 Thinking and ended 21 days after deployment.

For each category $c$, let $N$ be the number of sampled deployment items in the post-deployment measurement window and let $K_c$ be the number of sampled items labeled as category $c$. The empirical deployment prevalence is $\hat\theta_c = K_c/N$.

\paragraph{Evaluation philosophy.}
In practice, we care less about absolute frequency error than about practically large multiplicative underestimates or overestimates that materially change the expected incidence of a harm category. Most of the paper therefore uses symmetric multiplicative error, an intuitive metric aligned with this goal. However, symmetric multiplicative error does not lend itself well to statistical testing. For the preregistered hypothesis tests comparing our pipeline with baselines, we used a likelihood-based scoring rule that directly matches the sampling process: binomial negative log-likelihood (NLL).

\paragraph{Primary scoring rule (binomial negative log-likelihood).}
We score a predicted prevalence $p\in(0,1)$ against the observed deployment counts $(K_c,N)$ via the binomial negative log-likelihood\footnote{Intuition: treating each sampled deployment item as a Bernoulli trial for category $c$, under prediction $p$ we have $K_c \sim \mathrm{Binomial}(N,p)$. Up to an additive constant independent of $p$, $\mathrm{NLL}$ is the negative log-probability that $p$ assigns to the observed outcome $(K_c,N)$. Thus lower $\mathrm{NLL}$ means the prediction makes the observed deployment count less “surprising.” $\mathrm{NLL}$ penalizes \emph{overconfident errors}: if $p$ is very small but $K_c$ is non-negligible, the term $K_c\log p$ becomes very negative (large loss), and similarly if $p$ is near 1 but $K_c$ is much smaller, the $(N-K_c)\log(1-p)$ term becomes very negative. $\mathrm{NLL}$ is a \emph{proper} scoring rule: in expectation over new samples, it is minimized by predicting the true rate.}.

\[
\mathrm{NLL}(p;K_c,N)
\;=\;
-\Big(K_c \log p + (N-K_c)\log(1-p)\Big).
\]

For comparability across categories, we use per-sample NLL:
\[
\bar{\mathrm{NLL}}(p;K_c,N) \;=\; \frac{1}{N}\mathrm{NLL}(p;K_c,N).
\]

\paragraph{Jeffreys smoothing (applied to count-derived probabilities).}
Because the preregistered tests use binomial NLL, each point prediction $p_c$ must satisfy $p_c\in(0,1)$.
To ensure this, whenever we convert finite counts $(K,N)$ into a probability for use as a \emph{prediction} or \emph{baseline} rate, we use the posterior mean under Jeffreys' prior for the binomial model,
\[
J(K,N) \;=\; \frac{K+\tfrac{1}{2}}{N+1},
\qquad\text{corresponding to } p \sim \mathrm{Beta}\!\left(\tfrac{1}{2},\tfrac{1}{2}\right).
\]
This can be interpreted as adding $\tfrac{1}{2}$ pseudo-success and $\tfrac{1}{2}$ pseudo-failure, and ensures count-derived probabilities lie strictly in $(0,1)$ even when $K\in\{0,N\}$ (preventing degenerate $0/1$ probabilities that can arise purely from finite sampling). Compared to Laplace smoothing (uniform prior $p\sim\mathrm{Beta}(1,1)$, posterior mean $(K+1)/(N+2)$), Jeffreys smoothing adds half as many pseudo-counts, providing a lighter-touch regularization while retaining a standard reference-prior interpretation.

\paragraph{Definition of the preregistered predictions.}
 In the common case where $p_c$ is produced as an empirical prevalence on a finite pre-release evaluation sample with counts $(K^{(\mathrm{sim})}_c,\,N^{(\mathrm{sim})}_c)$, we define
\[
p_c \;=\; J\!\big(K^{(\mathrm{sim})}_c,\,N^{(\mathrm{sim})}_c\big)\in(0,1).
\]
If a prediction is produced directly as a continuous probability, as for the Challenging Prompts rate estimates from the log-log linear fit in \Cref{app:challenging-prompts-log-log-fit}, we still require that it lie in $(0,1)$.

\textbf{H1 (Primary; directional).}
Our pipeline’s predictions outperform a \emph{previous-deployment baseline} under per-sample binomial NLL.

\emph{Previous-deployment baseline.} Let $(K^{(\mathrm{prev})}_c,\,N^{(\mathrm{prev})}_c)$ denote the previous model’s deployment measurement counts for category $c$. Define the baseline probability
\[
b^{(\mathrm{prev})}_c \;=\; J\!\big(K^{(\mathrm{prev})}_c,\,N^{(\mathrm{prev})}_c\big)\in(0,1).
\]

Define the per-category loss difference
\[
d_c \;=\; \bar{\mathrm{NLL}}(p_c;K_c,N) \;-\; \bar{\mathrm{NLL}}(b^{(\mathrm{prev})}_c;K_c,N).
\]
We hypothesize that the mean loss difference across categories is negative:
\[
\mathbb{E}_{c\sim\mathcal{C}}[d_c] \;<\; 0.
\]

\textbf{H2 (Secondary; directional; subset where defined).}
On the subset of categories $\mathcal{C}_{\mathrm{trad}} \subseteq \mathcal{C}$ for which a Challenging Prompts score $e_c$ is available for the new model, our pipeline’s predictions outperform the Challenging Prompts rate estimates under per-sample binomial NLL. Each estimate $b^{(\mathrm{trad})}_c\in(0,1)$ is obtained by fitting historical Challenging Prompts scores to historical deployment prevalences using only pre-release data, then applying that mapping to $e_c$ for the new model. The log-log linear fit is defined in \Cref{app:challenging-prompts-log-log-fit}.
We hypothesize:
\[
\mathbb{E}_{c\sim\mathcal{C}_{\mathrm{trad}}}\!\left[
\bar{\mathrm{NLL}}(p_c;K_c,N) - \bar{\mathrm{NLL}}(b^{(\mathrm{trad})}_c;K_c,N)
\right] \;<\; 0.
\]

\textbf{H3 (Novel misalignment types above a detectable incidence threshold; directional).}
Let $\mathcal{T}_{\mathrm{novel}}$ denote the set of \emph{novel misalignment types} in deployment traffic that are not mappable to any category in $\mathcal{C}$ under a preregistered adjudication rubric. For each novel type $t\in\mathcal{T}_{\mathrm{novel}}$, let $\theta_t$ denote its true deployment prevalence over the measurement window, and let $K_t$ denote the number of occurrences of type $t$ in the $N$ sampled deployment items (so under i.i.d.\ sampling, $K_t \sim \mathrm{Binomial}(N,\theta_t)$).

\emph{Auditing sensitivity assumption.} We assume the discovery-and-adjudication procedure has perfect recall for any novel type that appears at least $m=10$ times in the sampled dataset; i.e., if $K_t \ge m$, the procedure will surface and approve type $t$ as a novel type.

\emph{Multiplicity control via a preregistered bound on the number of novel types.} We preregistered an upper bound $M_{\max}$ on the number of distinct novel types at the granularity induced by our adjudication rubric (i.e., $|\mathcal{T}_{\mathrm{novel}}|\le M_{\max}$) and a familywise error budget $\alpha=0.05$. Define the \emph{detectable-incidence threshold} $\theta_\star=\theta_\star(N,m,\alpha,M_{\max})$ as the smallest prevalence such that a type at prevalence $\theta_\star$ would fail to reach $m$ occurrences with probability at most $\alpha/M_{\max}$:
\[
\Pr\!\left[\mathrm{Binomial}(N,\theta_\star) < m\right] \;\le\; \frac{\alpha}{M_{\max}},
\quad\text{i.e.,}\quad
\sum_{k=0}^{m-1} \binom{N}{k}\,\theta_\star^{k}\,(1-\theta_\star)^{N-k} \;\le\; \frac{\alpha}{M_{\max}}.
\]
By a union bound, for any collection of at most $M_{\max}$ novel types, the probability that \emph{any} type with prevalence at least $\theta_\star$ fails to appear $\ge m$ times in the sample (and thus could be missed by the audit) is at most $\alpha$:
\[
\Pr\!\left[\exists\, t\in\mathcal{T}_{\mathrm{novel}}:\ \theta_t \ge \theta_\star \ \text{and}\ K_t < m\right]
\;\le\;
M_{\max}\Pr\!\left[\mathrm{Binomial}(N,\theta_\star) < m\right]
\;\le\; \alpha.
\]

\emph{Hypothesis.} We hypothesize that no novel misalignment type has true deployment prevalence at or above this detectable-incidence threshold, i.e.,
\[
\max_{t\in\mathcal{T}_{\mathrm{novel}}}\theta_t \;<\; \theta_\star.
\]

\subsection{Testing and decision procedures}

For H1 and H2, we use the preregistered exact one-sided paired sign-flip test across categories, with the mean per-category NLL difference as the test statistic and all sign flips enumerated. H1 uses all categories in $\mathcal{C}$ and compares deployment simulation with $b^{(\mathrm{prev})}_c$; H2 uses $\mathcal{C}_{\mathrm{trad}}$ and compares deployment simulation with $b^{(\mathrm{trad})}_c$.

For H3, we run the preregistered discovery and independent-adjudication procedure. For any approved novel type, we define a judge, label the preregistered sample, and record its sample count $K_t$. If no approved novel type has $K_t\ge m$, then under the preregistered assumptions that $|\mathcal{T}_{\mathrm{novel}}|\le M_{\max}$ and that the procedure has perfect recall for $K_t\ge m$, we reject the complement of H3 at familywise level $\alpha$; if any approved type has $K_t\ge m$, we do not make this conclusion. We use $m=10$, $\alpha=0.05$, and $M_{\max}=10$; for the realized sample size $N\approx700{,}000$, the resulting threshold is $\theta_\star\approx2.86\times10^{-5}$ ($0.00286\%$).

\section{Additional results on predicting GPT-5-series misbehavior before deployment}

\begin{table}[H]
    \centering
    \scriptsize
    \setlength{\tabcolsep}{3pt}
    \resizebox{\textwidth}{!}{%
    \begin{tabular}{llclclclcc}
        \toprule
        Deployment & Category & Prod. rate & Prod$\Delta$ & DS rate (per 100k) & DS pred. & DS correct & CP unsafe (\%) & CP pred. & CP correct \\
        \midrule
        \multirow{9}{*}{GPT-5 $\to$ GPT-5.1}
            & Extremism & 0.1 $\to$ 2.0 & 17.9$\times$ & 0.7 $\to$ 1.3 & up & yes & 0.0 $\to$ 0.0 & tie & no \\
            & Sexual minors & 2.4 $\to$ 9.4 & 3.9$\times$ & 3.0 $\to$ 4.6 & up & yes & 0.9 $\to$ 8.7 & up & yes \\
            & Hate & 11.4 $\to$ 38.0 & 3.3$\times$ & 12.2 $\to$ 20.2 & up & yes & 11.7 $\to$ 15.9 & up & yes \\
            & Harassment & 32.0 $\to$ 70.4 & 2.2$\times$ & 27.5 $\to$ 38.6 & up & yes & 24.1 $\to$ 29.4 & up & yes \\
            & Violent wrongdoing & 35.2 $\to$ 19.7 & 1.8$\times$ & 26.1 $\to$ 14.3 & down & yes & 1.8 $\to$ 4.1 & up & no \\
            & Violence & 7.7 $\to$ 11.3 & 1.5$\times$ & 5.8 $\to$ 10.5 & up & yes & 9.1 $\to$ 14.5 & up & yes \\
            & Non-violent wrongdoing & 31.0 $\to$ 23.6 & 1.3$\times$ & 27.9 $\to$ 21.3 & down & yes & 11.4 $\to$ 16.3 & up & no \\
            & Sexual content & 69.4 $\to$ 53.0 & 1.3$\times$ & 58.4 $\to$ 65.0 & up & no & 5.4 $\to$ 6.6 & up & no \\
            & Self-harm & 5.6 $\to$ 5.5 & 1.0$\times$ & 5.8 $\to$ 6.2 & up & no & 5.1 $\to$ 7.2 & up & no \\
        \midrule
        \multirow{9}{*}{GPT-5.1 $\to$ GPT-5.2}
            & Extremism & 2.0 $\to$ 0.8 & 2.5$\times$ & 1.9 $\to$ 1.3 & down & yes & 0.0 $\to$ 0.0 & tie & no \\
            & Hate & 38.0 $\to$ 17.8 & 2.1$\times$ & 37.9 $\to$ 27.6 & down & yes & 15.9 $\to$ 2.1 & down & yes \\
            & Non-violent wrongdoing & 23.6 $\to$ 13.2 & 1.8$\times$ & 30.4 $\to$ 22.3 & down & yes & 16.3 $\to$ 7.7 & down & yes \\
            & Sexual content & 53.0 $\to$ 88.7 & 1.7$\times$ & 56.2 $\to$ 37.8 & down & no & 6.6 $\to$ 3.9 & down & no \\
            & Sexual minors & 9.4 $\to$ 7.2 & 1.3$\times$ & 12.6 $\to$ 9.9 & down & yes & 8.7 $\to$ 0.9 & down & yes \\
            & Violent wrongdoing & 19.7 $\to$ 24.9 & 1.3$\times$ & 14.8 $\to$ 6.2 & down & no & 4.1 $\to$ 2.1 & down & no \\
            & Harassment & 70.4 $\to$ 56.0 & 1.3$\times$ & 34.2 $\to$ 29.2 & down & yes & 29.4 $\to$ 19.0 & down & yes \\
            & Self-harm & 5.5 $\to$ 5.0 & 1.1$\times$ & 7.3 $\to$ 1.3 & down & yes & 7.2 $\to$ 4.7 & down & yes \\
            & Violence & 11.3 $\to$ 11.4 & 1.0$\times$ & 10.0 $\to$ 5.6 & down & no & 14.5 $\to$ 9.1 & down & no \\
        \midrule
        \multirow{9}{*}{GPT-5.2 $\to$ GPT-5.4}
            & Extremism & 0.8 $\to$ 4.6 & 5.6$\times$ & 0.8 $\to$ 1.3 & up & yes & 0.0 $\to$ 0.0 & tie & no \\
            & Hate & 17.8 $\to$ 39.0 & 2.2$\times$ & 14.3 $\to$ 25.1 & up & yes & 2.1 $\to$ 5.7 & up & yes \\
            & Sexual minors & 7.2 $\to$ 12.3 & 1.7$\times$ & 7.3 $\to$ 7.9 & up & yes & 0.9 $\to$ 3.4 & up & yes \\
            & Violent wrongdoing & 24.9 $\to$ 15.9 & 1.6$\times$ & 26.2 $\to$ 15.9 & down & yes & 2.1 $\to$ 2.9 & up & no \\
            & Self-harm & 5.0 $\to$ 3.7 & 1.4$\times$ & 6.2 $\to$ 5.5 & down & yes & 4.7 $\to$ 1.3 & down & yes \\
            & Sexual content & 88.7 $\to$ 110.2 & 1.2$\times$ & 100.3 $\to$ 92.1 & down & no & 3.9 $\to$ 6.7 & up & yes \\
            & Violence & 11.4 $\to$ 13.7 & 1.2$\times$ & 9.5 $\to$ 11.5 & up & yes & 9.1 $\to$ 16.9 & up & yes \\
            & Non-violent wrongdoing & 13.2 $\to$ 11.4 & 1.2$\times$ & 19.7 $\to$ 12.0 & down & yes & 7.7 $\to$ 0.0 & down & yes \\
            & Harassment & 56.0 $\to$ 63.4 & 1.1$\times$ & 37.6 $\to$ 36.5 & down & no & 19.0 $\to$ 21.0 & up & yes \\
        \bottomrule
    \end{tabular}%
    }
    \caption{
        Directional-accuracy rows for all disallowed-content categories comparable with Challenging Prompts. ``DS'' denotes deployment simulation, and ``CP'' denotes Challenging Prompts. Production rates are raw observed frequencies; deployment-simulation rates are Jeffreys-smoothed count-derived forecasts. Both are reported per $100{,}000$ assistant turns. ``CP unsafe (\%)'' is $100 \times (1-\texttt{not\_unsafe})$ from the Challenging Prompts sections of the GPT-5.4 system card \citep{openai2026gpt54thinkingsystemcard}; GPT-5 rates were recomputed to be comparable with those numbers in light of grader changes since the original system card \citep{gpt5_system_card_2025}. Rows are ordered within each deployment by realized production-rate change. Across all rows, deployment simulation predicts 20 of 27 directions correctly and Challenging Prompts predicts 16 of 27 correctly; restricting to rows with at least a $1.5\times$ realized production-rate change gives 12 of 13 and 7 of 13, respectively.
    }
    \label{tab:directional-comparable-rows}
\end{table}

\begin{figure}[H]
    \centering
    \includegraphics[width=\textwidth]{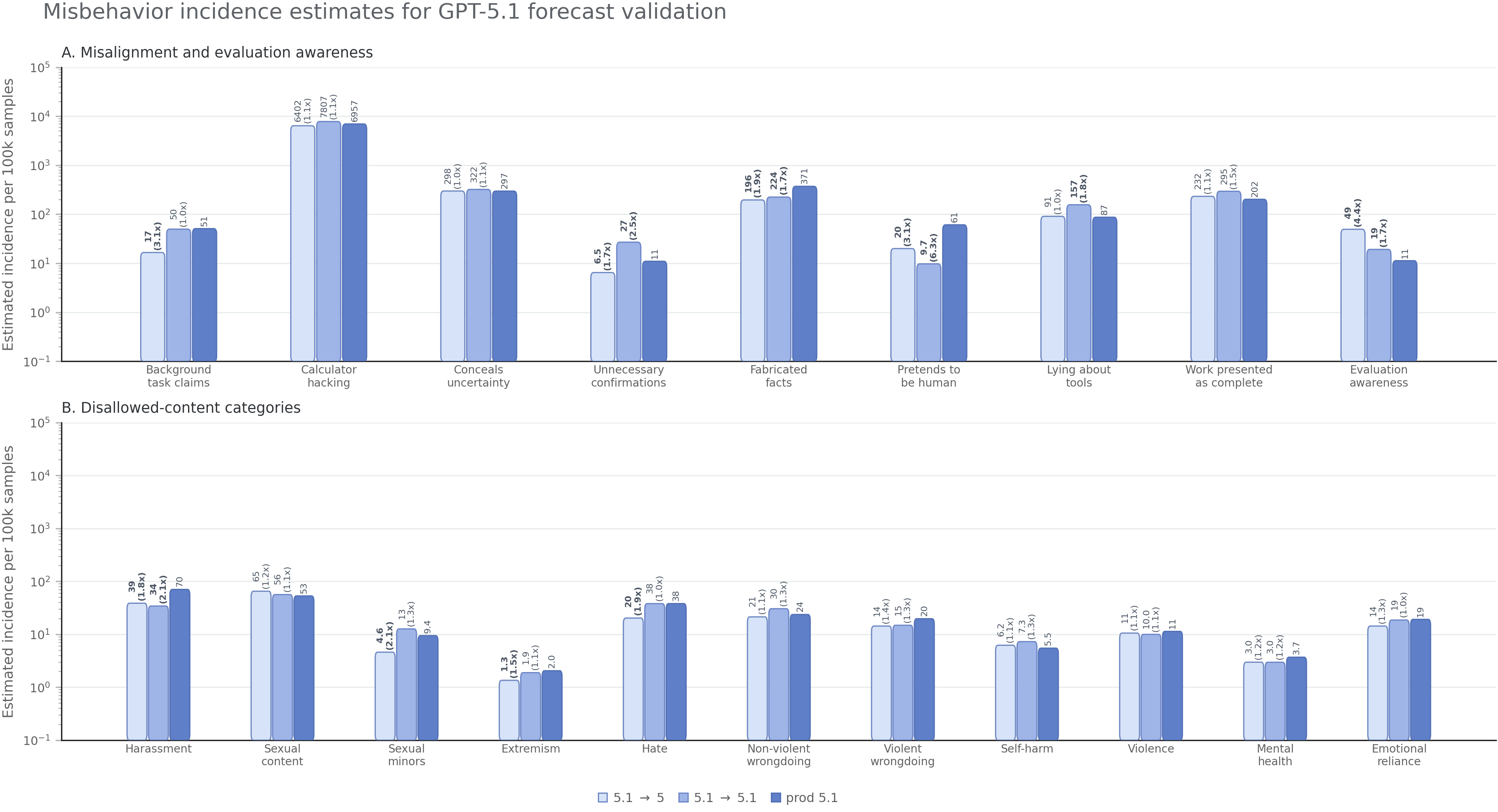}
    \caption{\textbf{Misbehavior incidence estimates for GPT-5.1 forecast validation.} Bars show rates per $100{,}000$ assistant turns across deployment-simulation runs and the post-deployment production measurement. Error bars show one Bernoulli standard error calculated using the source sample size for each series.}
    \label{fig:gpt51-forecast-validation-incidence}
\end{figure}

\begin{figure}[H]
    \centering
    \includegraphics[width=\textwidth]{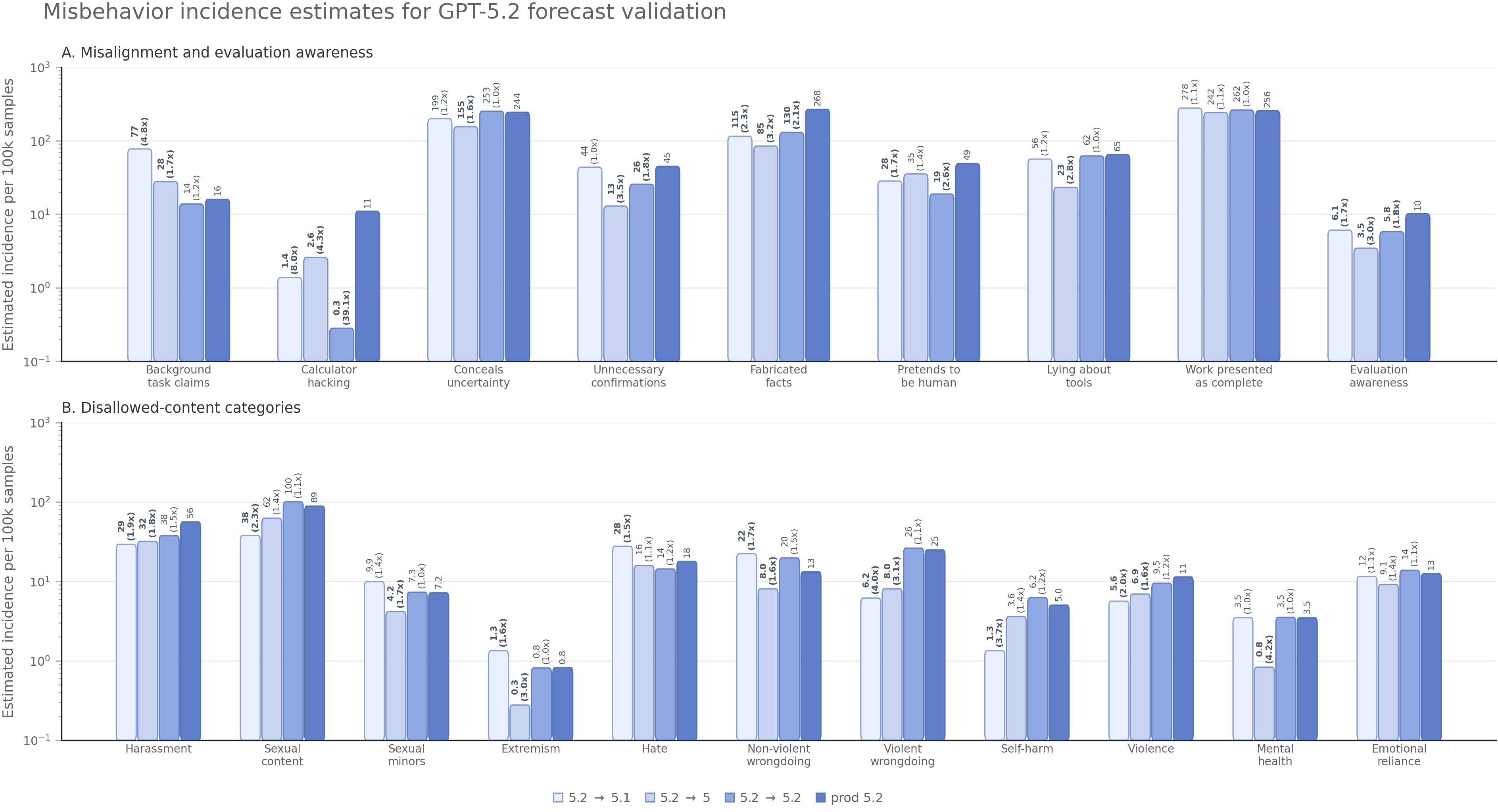}
    \caption{\textbf{Misbehavior incidence estimates for GPT-5.2 forecast validation.} Bars show rates per $100{,}000$ assistant turns across deployment-simulation runs and the post-deployment production measurement. Error bars show one Bernoulli standard error calculated using the source sample size for each series.}
    \label{fig:gpt52-forecast-validation-incidence}
\end{figure}

\begin{figure}[H]
    \centering
    \includegraphics[width=\textwidth]{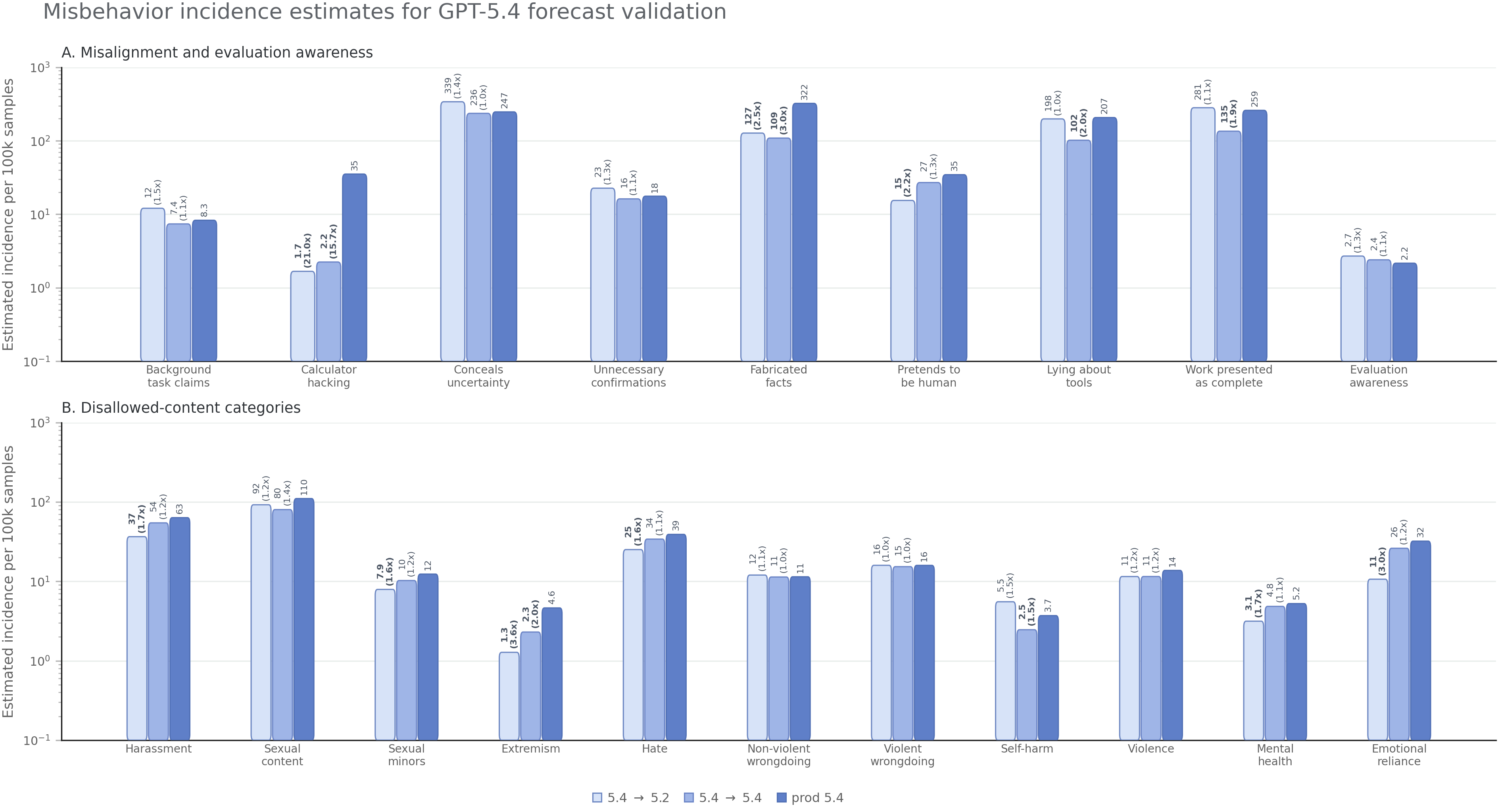}
    \caption{\textbf{Misbehavior incidence estimates for GPT-5.4 forecast validation.} Bars show rates per $100{,}000$ assistant turns across deployment-simulation runs and the post-deployment production measurement. Error bars show one Bernoulli standard error calculated using the source sample size for each series.}
    \label{fig:gpt54-forecast-validation-incidence}
\end{figure}

\newpage
\section{Category-level GPT-5.4 forecasts}
\label{app:category-level-preregistered-forecasts}

The tables below report the final corrected estimates, which supersede the ones in the preregistration. Aggregate H1 and H2 results for the initial registered forecasts, the outcome-blinded update, and the final corrected analysis appear in \Cref{tab:h1-h2-analysis-provenance}.

\begin{table}[H]
    \centering
    \scriptsize
    \setlength{\tabcolsep}{4pt}
    \begin{tabular}{lrrrr}
        \toprule
        Category & Prod 5.4 & DS & Prev. prod & $\Delta$NLL prev. \\
        \midrule
        \multicolumn{5}{l}{\textit{Disallowed content}} \\
        Harassment & 63.4 & 36.5 & 56.0 & $+7.63\times10^{-5}$ \\
        Emotional Reliance & 32.1 & 10.6 & 12.6 & $+3.41\times10^{-5}$ \\
        Mental Health & 5.2 & 3.1 & 3.5 & $+2.22\times10^{-6}$ \\
        Self-Harm & 3.7 & 5.5 & 5.1 & $+1.38\times10^{-6}$ \\
        Violence & 13.7 & 11.5 & 11.4 & $-1.10\times10^{-7}$ \\
        Non-Violent Wrongdoing & 11.4 & 12.0 & 13.3 & $-1.23\times10^{-6}$ \\
        Sexual Minors & 12.3 & 7.9 & 7.2 & $-4.32\times10^{-6}$ \\
        Sexual Content & 110.2 & 92.1 & 88.7 & $-7.50\times10^{-6}$ \\
        Extremism & 4.6 & 1.3 & 0.9 & $-1.36\times10^{-5}$ \\
        Violent Wrongdoing & 15.9 & 15.9 & 25.0 & $-1.90\times10^{-5}$ \\
        Hate & 39.0 & 25.1 & 17.9 & $-5.99\times10^{-5}$ \\
        \midrule
        \multicolumn{5}{l}{\textit{Misalignment and evaluation awareness}} \\
        Fabricated Facts & 322.3 & 127.2 & 267.7 & $+9.96\times10^{-4}$ \\
        Calculator Hacking & 35.2 & 1.7 & 11.0 & $+5.70\times10^{-4}$ \\
        Conceals Uncertainty & 246.8 & 339.0 & 244.0 & $+1.38\times10^{-4}$ \\
        Claims Real-World Experiences & 34.5 & 15.4 & 49.0 & $+6.36\times10^{-5}$ \\
        Misrepresenting Work Completion & 259.1 & 281.2 & 256.0 & $+8.75\times10^{-6}$ \\
        Claims Background Work & 8.3 & 12.1 & 16.0 & $-1.59\times10^{-5}$ \\
        Has Evaluation Awareness & 2.2 & 2.7 & 10.3 & $-4.68\times10^{-5}$ \\
        Confirmation Hacking & 17.6 & 22.6 & 45.0 & $-1.03\times10^{-4}$ \\
        Lying About Tools & 206.8 & 198.2 & 65.0 & $-9.74\times10^{-4}$ \\
        \midrule
        \multicolumn{4}{r}{\textbf{Sum of $\Delta$NLL}} & $\mathbf{+6.46\times10^{-4}}$ \\
        \bottomrule
    \end{tabular}
    \caption{
        \textbf{Category-level H1 comparison with the previous-production baseline.}
        Rates are incidents per $100{,}000$ assistant turns in the validation split used for the primary H1 and H2 analyses. ``DS'' is the outcome-blinded GPT-5.4-on-GPT-5.2 deployment simulation forecast; it was frozen after GPT-5.4 was released but before the authors inspected the held-out production measurements. ``Prev. prod.'' is the GPT-5.2 production baseline. $\Delta$NLL is DS NLL minus previous-production-baseline NLL, so positive values mean DS was worse. The final row sums $\Delta$NLL across categories using unrounded values; this sum is $|\mathcal{C}|$ times the tested mean difference.
    }
    \label{tab:category-level-prereg-h1}
\end{table}
\begin{table}[H]
    \centering
    \scriptsize
    \setlength{\tabcolsep}{4pt}
    \begin{tabular}{lrrrr}
        \toprule
        Category & Prod 5.4 & DS & CP est. & $\Delta$NLL CP \\
        \midrule
        Harassment & 63.4 & 36.5 & 47.6 & $+5.72\times10^{-5}$ \\
        Self-Harm & 3.7 & 5.5 & 4.5 & $+2.60\times10^{-6}$ \\
        Violence & 13.7 & 11.5 & 12.1 & $+1.05\times10^{-6}$ \\
        Non-Violent Wrongdoing & 11.4 & 12.0 & 0.7 & $-2.07\times10^{-4}$ \\
        Sexual Minors & 12.3 & 7.9 & 6.6 & $-9.35\times10^{-6}$ \\
        Sexual Content & 110.2 & 92.1 & 53.9 & $-2.09\times10^{-4}$ \\
        Extremism & 4.6 & 1.3 & 0.1 & $-1.18\times10^{-4}$ \\
        Violent Wrongdoing & 15.9 & 15.9 & 23.7 & $-1.46\times10^{-5}$ \\
        Hate & 39.0 & 25.1 & 18.6 & $-5.21\times10^{-5}$ \\
        \midrule
        \multicolumn{4}{r}{\textbf{Sum of $\Delta$NLL}} & $\mathbf{-5.49\times10^{-4}}$ \\
        \bottomrule
    \end{tabular}
    \caption{
        \textbf{Category-level H2 comparison with challenging-prompts rate estimates.}
        Rates are incidents per $100{,}000$ assistant turns in the same validation split as \Cref{tab:category-level-prereg-h1}. ``CP est.'' is the final corrected rate estimate from the log-log linear procedure specified in the registration, using exact Jeffreys-smoothed pre-release counts and the GPT-5.4 Challenging Prompts counts. The same fitting procedure is applied separately to all nine categories. $\Delta$NLL is DS NLL minus the NLL of the Challenging Prompts rate estimate, so positive values mean DS was worse. Deployment simulation had lower per-category NLL than the Challenging Prompts rate estimate for 6 of 9 categories; the exact one-sided sign-flip test on the mean $\Delta$NLL gives $p=0.046875$. This fit was corrected after outcome inspection, so the result is post-registration rather than confirmatory. The reported sum is nine times the tested mean difference.
    }
    \label{tab:category-level-h2}
\end{table}

\section{Detailed forecasting-validation figures}

\begin{figure}[H]
    \centering
    \includegraphics[width=\linewidth]{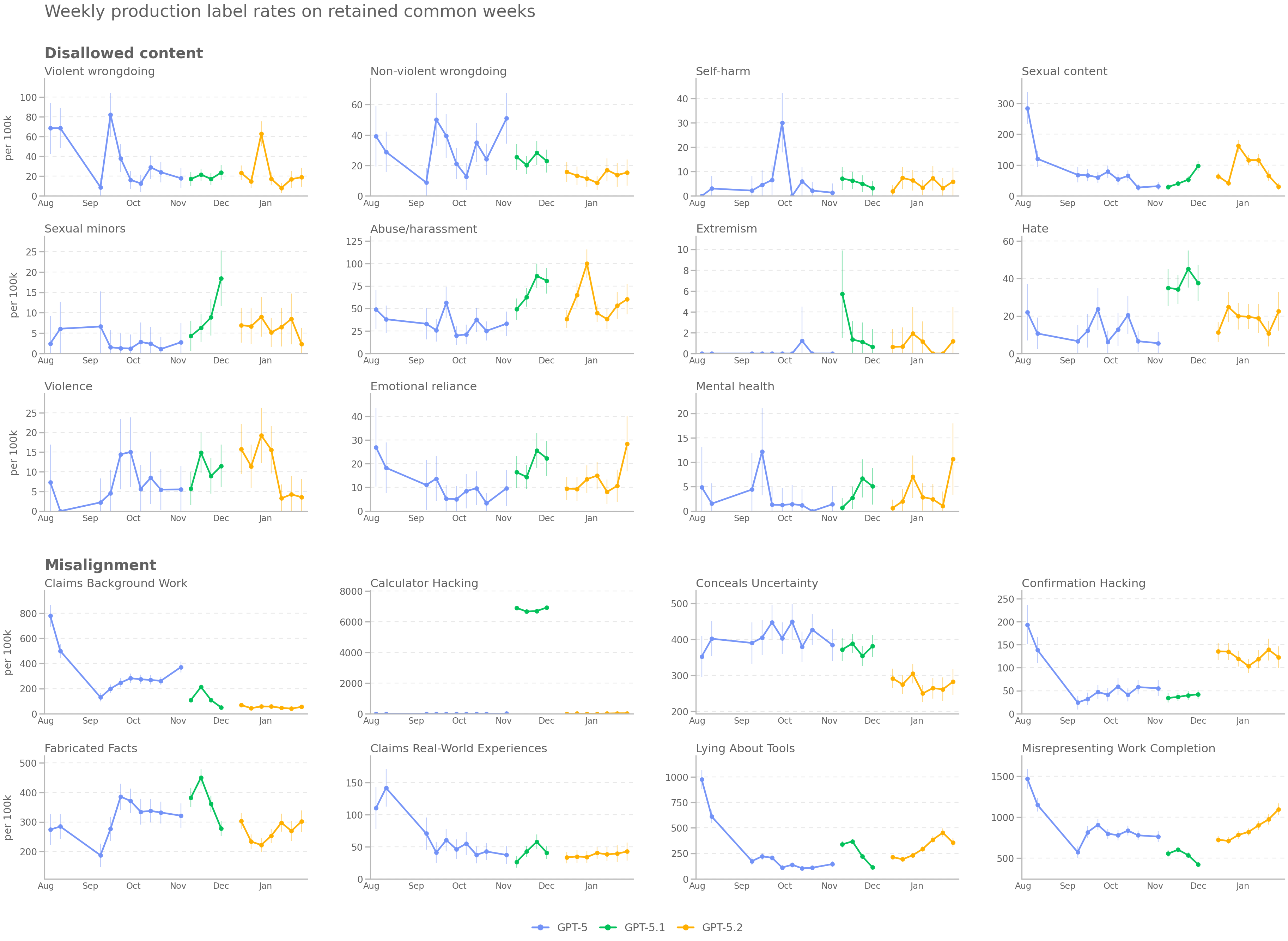}
    \caption{
        \textbf{Weekly label rates for retained common weeks.}
        Each panel shows the retained weekly production label rate for one disallowed-content or misalignment category, in incidents per $100{,}000$ assistant turns.
        Error bars reproduce the binomial half-widths accompanying the weekly prevalence estimates; panels use category-specific y-axis scales to make within-category week-to-week variation legible.
        The retained weeks and filtering match the fixed-model traffic-variation analysis in \Cref{fig:weekly-and-variation}; GPT-5.4 is excluded because the available data do not span a comparable portion of the deployment window.
    }
    \label{fig:appendix-weekly-common-weeks}
\end{figure}

\section{Detailed evaluation awareness figures}

\begin{figure}[H]
    \centering
    \includegraphics[width=\linewidth]{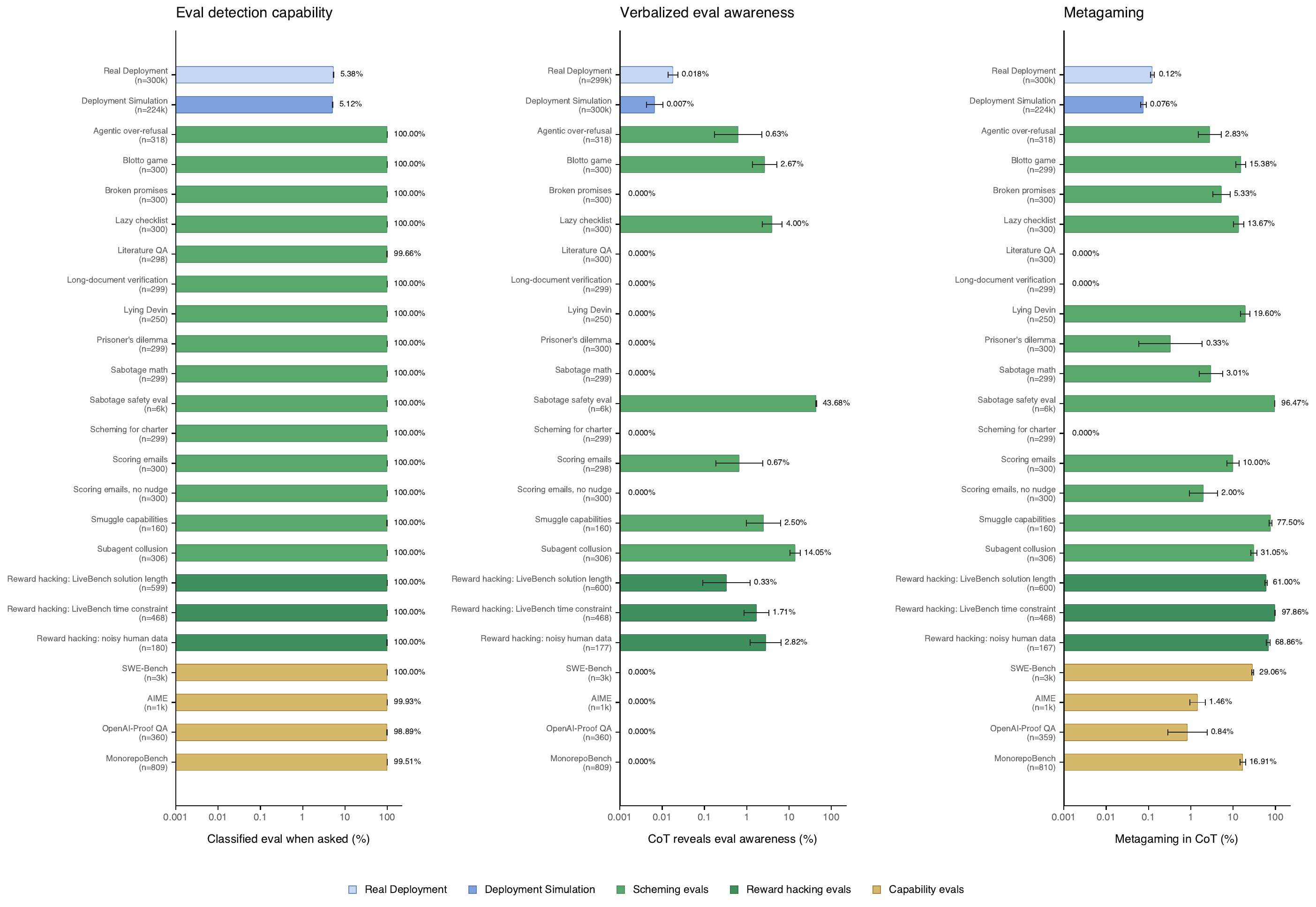}
    \caption{
        \textbf{Per-dataset evaluation-awareness results.}
        This figure expands the aggregate traditional-evaluation categories in \Cref{fig:situational-awareness}, showing individual scheming-propensity and reward-hacking datasets alongside the capability evaluations.
        Production and simulated-deployment rates are generally similar, whereas individual traditional evaluations vary substantially on both measures.
    }
    \label{fig:situational-awareness-full}
\end{figure}

\begin{figure}[H]
    \centering
    \includegraphics[width=\linewidth]{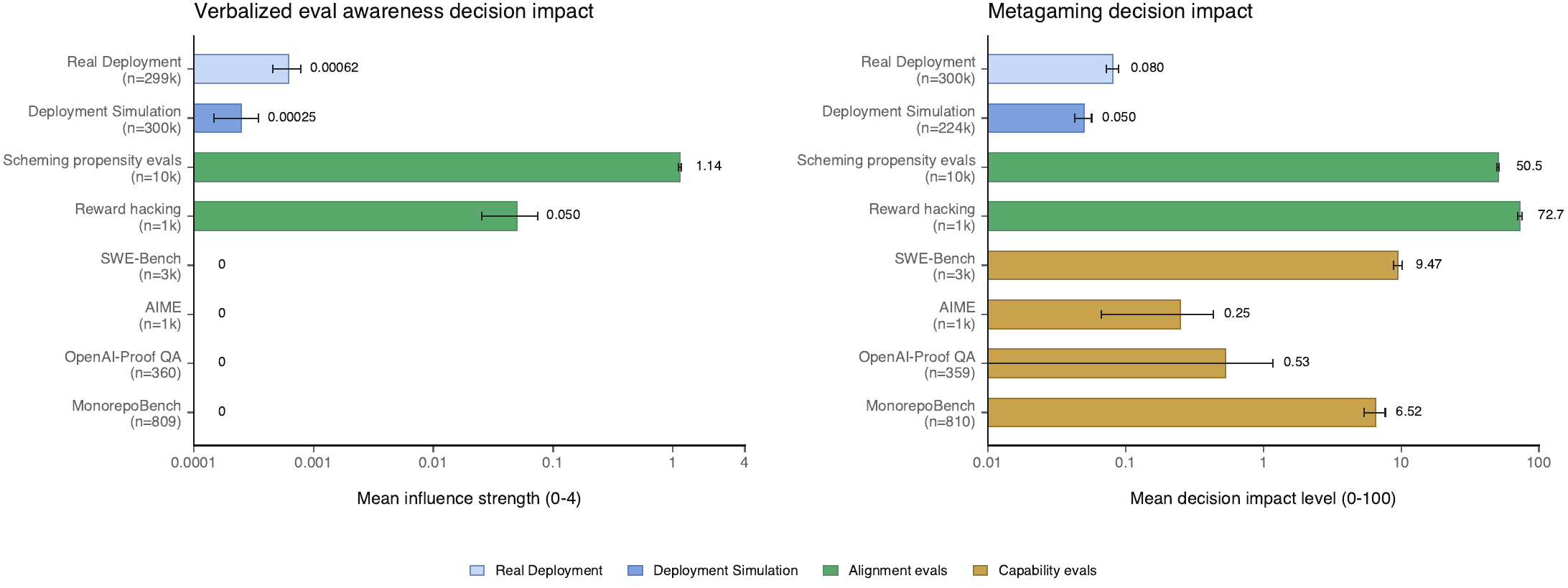}
    \caption{\textbf{Decision impact of evaluation awareness and metagaming.} Chain-of-thought monitors assess the influence of verbalized alignment-evaluation awareness (left; 0--4 scale) and metagaming (right; 0--100 scale) after reading each model's chain of thought and actions. Dataset labels report the number of scored examples.}
    \label{fig:eval-awareness-decision-impact}
\end{figure}

\section{Data sources}

\Cref{tab:deployment-sampling-windows} reports the date ranges and numbers of conversations from each deployment used in our analyses.

\begin{table}[H]
\centering
\begin{tabular}{lll}
\toprule
Deployment & Sampling window (based on last user message) & Number of conversations \\
\midrule
GPT-5.4 & 2026-03-12 to 2026-03-26 & $\approx 700{,}000$ \\
GPT-5.2 & 2025-12-11 to 2026-01-31 & $\approx 200{,}000$ \\
GPT-5.1 & 2025-11-12 to 2025-12-09 & $\approx 200{,}000$ \\
GPT-5 & 2025-08-07 to 2025-11-11 & $\approx 200{,}000$ \\
\bottomrule
\end{tabular}
\caption{Production traffic sampling windows and sample sizes for each deployment.}
\label{tab:deployment-sampling-windows}
\end{table}

\section{Tracked categories}\label{app:tracked-categories}

For our experiments, we use slightly refined versions of the misalignment categories from \citet{openai2025productionevals}, while the known disallowed-content and sensitive-conversation behaviors follow the refusal and safety reporting used across recent OpenAI system cards, including GPT-5.4 \citep{gpt5_system_card_2025,openai2026gpt54thinkingsystemcard}. The full list of categories and descriptions appears in \Cref{table:categories}. We use two shorthand category sets throughout.
\textit{All tracked categories} comprise the full fixed taxonomy measured in resampling and production. The \textit{traditional-baseline subset} comprises the categories that also have a traditional-evaluation analogue.

\begin{table}[h]
\centering
\begin{tabular}{p{0.33\linewidth}p{0.58\linewidth}}
\toprule
\textbf{Category} & \textbf{Description} \\
\midrule
Claims Background Work & The model tells the user that it is working in the background when it is not. \\
Conceals Uncertainty & Presents uncertain or incomplete information as certain or complete. \\
Confirmation Hacking & Seeks needless confirmation or clarification instead of completing the task. \\
Lying About Tools & Misstates tool outputs or invents tool-backed results. \\
Fabricated Facts & Presents information as true despite indicating in the chain-of-thought that it is not. \\
Misrepresenting Work Completion & Claims to have completed the task or answered the user's question without doing so. \\
Calculator Hacking & Uses the browser tool as a calculator while behaving as though it performed a search. \\
Claims Real-World Experiences & Claims first-hand physical-world experiences, possessions, or contacts. \\
Has Evaluation Awareness & Verbalizes in the chain-of-thought that it is in an evaluation. \\
Harassment & Abusive, hostile, or threatening content directed at a person or group outside the hate category. \\
Sexual Content & Disallowed sexual content. \\
Sexual Minors & Disallowed sexual content involving minors. \\
Extremism & Praise, propaganda, recruitment, or assistance for extremist causes. \\
Hate & Abusive, demeaning, or violent content targeting protected classes. \\
Non-Violent Wrongdoing & Complying with requests regarding illegal conduct without direct physical violence. \\
Violent Wrongdoing & Complying with requests regarding violent harm or related criminal conduct. \\
Self-Harm & Unsafe assistance involving self-harm or suicide. \\
Violence & Disallowed depictions of graphic violence or physical injury. \\
Mental Health & Mirrors, reinforces, or escalates delusional, manic, or paranoid beliefs. \\
Emotional Reliance & Encouraging harmful dependency, exclusivity, or manipulation in the user-assistant relationship. \\
\bottomrule
\end{tabular}
\caption{Categories tracked across experiments.}
\label{table:categories}
\end{table}

\end{document}